\documentclass[10pt,twocolumn,letterpaper]{article}

\usepackage{cvpr}              %

\usepackage{graphicx}
\usepackage{amsmath}
\usepackage{mathtools}
\usepackage{amssymb}
\usepackage{booktabs}
\usepackage{multicol, blindtext}
\usepackage{subcaption}
\usepackage{cite}
\usepackage{soul}
\usepackage{units}
\usepackage{array}
\usepackage{multirow}
\usepackage{caption}
\usepackage{float}
\usepackage{pifont}
\usepackage{changepage}

\makeatletter
\@namedef{ver@everyshi.sty}{}
\makeatother
\usepackage{tikz}
\usetikzlibrary{calc} 
\usepackage{pgfplots}
\usepackage{todonotes}
\usepackage{colortbl}
\usepackage{amssymb}


\usepackage[pagebackref,breaklinks,colorlinks]{hyperref}

\usepackage[capitalize]{cleveref}
\crefname{section}{Sec.}{Secs.}
\Crefname{section}{Section}{Sections}
\Crefname{table}{Table}{Tables}
\crefname{table}{Tab.}{Tabs.}

\newcommand{\C}{\mathbf{C}}
\newcommand{\ahat}{\mathbf{\hat{\boldsymbol{\alpha}}}}
\newcommand{\lhat}{\mathbf{\hat{\boldsymbol{\Lambda}}}}
\newcommand{\xhat}{\mathbf{\hat{\boldsymbol{X}}}}
\newcommand{\rhat}{\mathbf{\hat{r}}}
\newcommand{\Zhat}{\mathbf{\hat{Z}}}

\newcolumntype{L}[1]{>{\raggedright\arraybackslash}p{#1}}
\newcolumntype{C}[1]{>{\centering\arraybackslash}p{#1}}
\newcolumntype{R}[1]{>{\raggedleft\arraybackslash}p{#1}}

\newcommand{\Z}{\mathbf{Z}}
\definecolor{dai_petrol}{RGB}{0,103,127}
\definecolor{dai_deepred}{RGB}{112,23,11}
\definecolor{dai_gray}{RGB}{146,146,146}
\definecolor{dai_orange}{RGB}{255,127,42}
\definecolor{dai_green}{RGB}{84,127,42}

\colorlet{light-gray}{gray!40}
\def\cvprPaperID{5903} %
\def\confName{CVPR}
\def\confYear{2022}

\title{Gated2Gated: Self-Supervised Depth Estimation from Gated Images}

\author{Amanpreet Walia\footnote{Equal Contribution}$^{\;\;1,3}$\hspace{0.5cm}Stefanie Walz\footnotemark[\value{footnote}]$^{\;\;2}$\hspace{0.5cm}Mario Bijelic$^{4}$\hspace{0.5cm}Fahim Mannan$^{1}$
\\Frank Julca-Aguilar$^{1}$\hspace{0.5cm}Michael Langer$^{3}$\hspace{0.5cm}Werner Ritter$^{2}$\hspace{0.5cm}Felix Heide$^{1,4}$ \vspace{0.25cm}\\
$^{1}$Algolux\hspace{0.5cm}$^{2}$Mercedes-Benz AG\hspace{0.5cm}$^{3}$McGill University\hspace{0.5cm}$^{4}$Princeton University
}

\begin{document}

\twocolumn[{%
         \vspace{-7mm}
         \renewcommand\twocolumn[1][]{#1}%
         \maketitle
         \vspace{-3mm}
         \thispagestyle{empty}
      }]
\def\thefootnote{*}\footnotetext{These authors contributed equally to this work.} %
\def\thefootnote{\arabic{footnote}}
\begin{abstract}
   Gated cameras hold promise as an alternative to scanning LiDAR sensors with high-resolution 3D depth that is robust to back-scatter in fog, snow, and rain. Instead of sequentially scanning a scene and directly recording depth via the photon time-of-flight, as in pulsed LiDAR sensors, gated imagers encode depth in the relative intensity of a handful of gated slices, captured at megapixel resolution. Although existing methods have shown that it is possible to decode high-resolution depth from such measurements, these methods require synchronized and calibrated LiDAR to supervise the gated depth decoder -- prohibiting fast adoption across geographies, training on large unpaired datasets, and exploring alternative applications outside of automotive use cases. In this work, we fill this gap and propose an entirely self-supervised depth estimation method that uses gated intensity profiles and temporal consistency as a training signal. The proposed model is trained end-to-end from gated video sequences, does not require LiDAR or RGB data, and learns to estimate absolute depth values. We take gated slices as input and disentangle the estimation of the scene albedo, depth, and ambient light, which are then used to learn to reconstruct the input slices through a cyclic loss. We rely on temporal consistency between a given frame and neighboring gated slices to estimate depth in regions with shadows and reflections. We experimentally validate that the proposed approach outperforms existing supervised and self-supervised depth estimation methods based on monocular RGB and stereo images, as well as supervised methods based on gated images. Code is available at \href{https://github.com/princeton-computational-imaging/Gated2Gated}{\texttt{\small https://github.com/princeton-computational-\\imaging/Gated2Gated.}}

\end{abstract}

\section{Introduction}\label{sec:intro}
Depth sensing has become a cornerstone imaging modality for 3D scene understanding. Depth is used directly as input to a vision module, or indirectly in training datasets, to supervise models relying on other modalities across a wide range of applications such as perception and planning in autonomous driving, robotics, remote sensing, augmented and virtual reality~\cite{MovingPeopleMovingCameras,PSeudoLidar}.  

The most successful existing methods for depth estimation are either based on pulsed scanning LiDAR~\cite{ma2018sparse}, or passive RGB sensors, i.e., monocular~\cite{eigen2014depth,Godard2017,saxena2006learning} and stereo RGB~\cite{hartley2003multiple,scharstein2003high}. LiDAR depth sensors~\cite{schwarz2010lidar} measure the time-of-flight of pulses of light emitted into the scene and returned to the sensor along a coaxial path that is sequentially scanned across a scene. As a result, these sensors deliver precise depth with high spatial resolution at short distances. However, they are expensive, suffer from quadratic decreasing spatial resolution at longer distances, e.g., resulting in a few measurement points for pedestrians at 100~m distance~\cite{schwarz2010lidar}, and they fail in the presence of strong back-scatter. Monocular and stereo RGB methods offer a substantially cheaper alternative, but they struggle to achieve depth precision comparable to time-of-flight imaging, struggle in low-light scenarios, and at long distances that map to small disparities.

Recently, gated imaging has been proposed as an alternative sensor modality for depth estimation and 3D detection~\cite{gated2depth2019,gated3d2021} which promise to overcome the spatial resolution limitation of scanning LiDAR while providing comparable depth precision. Gated cameras combine low-cost CMOS sensors with analog gated readout and active flash illumination, allowing to capture a sequence of gated image slices that each encode time-resolved illumination via their relative intensity and, as such, provide depth cues not present in RGB cameras. Thanks to this active gated flash acquisition mode, gated imaging methods have shown to be more robust in low-light scenarios and in the presence of strong back-scatter that can be suppressed during acquisition~\cite{gated2depth2019}. Gated images provide dense depth information at megapixel resolution of the gated camera, allowing for long-range perception where LiDAR-based methods fail~\cite{gated3d2021}. However, all of these existing methods require calibrated and synchronized LiDAR data for training supervision. Although commercial gated cameras have become available~\cite{grauer2014active}, the requirement of such a multi-modal capture system prohibits the rapid adoption of gated depth imaging not only in automotive applications but also in other robotic use cases. Moreover, large unpaired sequences of gated imagery cannot be exploited in training existing gated depth estimation methods. 
In this work, we propose the first self-supervised method for depth estimation from gated cameras. The proposed method takes gated slices as input and predicts the scene albedo, depth, and ambient illumination. We reconstruct the input slices using calibrated gated profiles enforcing \emph{measurement cycle consistency} and warping from the nearby slices in a temporal window enforcing \emph{temporal consistency}. 
To this end, we introduce a differentiable gated image formation model that uses depth-dependent calibrated gating profiles for self-supervised measurement cycle loss.
To learn in the absence of reliable gated measurements due to shadows, we utilize temporal depth consistency using differentiable structure-from-motion and the proposed image formation model.

Specifically, we make the following main contributions: 
\begin{itemize}
	\itemsep-0.3em
  \item We propose a novel self-supervised method that uses measurement cycle consistency and temporal consistency as training signals. 
  \item The proposed model is trained end-to-end, and by exploiting calibrated gating profiles, the method is able to accurately estimate metric depth using the cycle consistency component.
  \item We validate that the proposed method outperforms self-supervised and supervised depth estimation using monocular and stereo RGB images and supervised gated depth estimation methods. 
\end{itemize}
We will release all models and code used to reproduce the results from this work.

\vspace{-0.12cm}
\section{Related Work}
\label{sec:formatting}

\vspace{-0.3eM}
\paragraph{Depth from Time-of-Flight }
Time-of-Flight (ToF) cameras acquire depth by measuring the round-trip time of modulated flood-illuminated light returned from a scene. Existing methods can be classified into three categories: correlation \cite{hansard2012time, kolb2010time, lange00tof}, pulsed ToF \cite{schwarz2010lidar} and gated imaging \cite{heckman1967,grauer2014active}.  
Correlation ToF cameras \cite{hansard2012time, kolb2010time,
lange00tof} estimate the depth from the phase difference of the sent and received laser pulse, which allows high spatial resolution, but is limited to short distances and indoor environments \cite{heide2015doppler}.
Pulsed ToF sensors \cite{schwarz2010lidar} emit pulses of light and directly measure the round-trip time of the returned pulses, but require a scanning mechanism for large distances reducing the spatial resolution. Additionally, experiments have shown that they can be affected by adverse weather disturbances \cite{BenchmarkLidar,LIBRE,Jokela}.  
Gated imaging \cite{heckman1967,grauer2014active,Bijelic2018} records the returned light within a short integration time on the imaging sensor. This limits the capture to certain depth ranges and allows short-range back-scatter to be ignored. In \cite{Busck2004, Busck2005, Andersson2006} a sequence of three gated slices was used to reconstruct depth information. Further methods introduced analytical approximations \cite{Laurenzis2007, Laurenzis2009, Xinwei2013} or learned the depth prediction through Bayesian methods \cite{adam2017bayesian,schober2017dynamic} and deep neural networks \cite{gruber2018learning}.

\vspace{-1.0eM}
\paragraph{Supervised Depth Estimation}
Learning to predict depth from intensity images requires appropriate ground truth data. Methods either use the supervision from time-of-flight data \cite{eigen2014depth, chang2018pyramid,jaritz2018sparse,ma2018sparse,gated2depth2019} or ground truth depth from multi-view systems \cite{MovingPeopleMovingCameras,Mayer2016,Kendall2017}.
Previous imaging systems either process images from monocular cameras \cite{MovingPeopleMovingCameras,Chen2018b, Laina2016} can reason about multiple views~\cite{chang2018pyramid,zhang2019ga} or utilize the combination of monocular images with sparse LiDAR pointclouds~\cite{jaritz2018sparse, ma2018sparse}.
All of these existing methods can fail in low-light or low-contrast scenarios, for example, at night or in cluttered scenes, that active methods~\cite{gated2depth2019} tackle using illumination. Furthermore, RGB-only monocular depth estimation can only reconstruct up to an unknown scale factor. 
\vspace{-1.eM}
\paragraph{Self-Supervised Depth Estimation}
Acquiring ground truth data for supervised depth estimation methods is challenging. An extensive process was applied in \cite{geiger2012we,Uhrig2017THREEDV} to overcome the limited range and spatial resolution by requiring a thorough LiDAR camera synchronization with the ego-motion correction and accumulating single point clouds into LiDAR maps as ground truth. 
Nonetheless, the application spectrum is limited, disallowing the use in areas such as scattering media, where LiDAR data is cluttered \cite{Bijelic_2020_STF} or for vehicle fleet data without expensive ground truth sensors.
To tackle this challenge, self-supervised training approaches exploit multiview geometry by aligning stereo image pairs~\cite{Garg2016, godard2017unsupervised} or making use of image view synthesis between temporally consecutive 
frames~\cite{Zhou2017, godard2019digging,guizilini20203d}.  
Aligning stereo images pairs for depth prediction was initially proposed by Garg et al. \cite{Garg2016}. Here, a neural network predicts the disparity from monocular camera images and supervises it by warping the stereo images. %
Image synthesis between temporally consecutive monocular image frames was introduced in \cite{Ummenhofer2017,godard2017unsupervised}. It utilizes two independent networks, one predicting the depth and the other estimating a rigid body transformation between two temporally adjacent frames. A reprojection error between two frames is then formulated to supervise the depth estimate. 
The following monocular methods investigate novel neural architectures \cite{Garg2016,godard2019digging,guizilini20203d} or extensions in the loss formulation \cite{godard2017unsupervised,vijayanarasimhan2017sfm, yin2018geonet, ranjan2019competitive, godard2019digging, luo2019every, dai2020self, guizilini20203d}. However, they have inherent scale ambiguity, which can be reduced by relying on vehicle velocity or LiDAR ground-truth depth measurements at test-time \cite{guizilini20203d}. 
Departing from these approaches, the proposed depth estimation method relies on the calibrated gate profiles used in a measurement cycle consistency loss to enforce scale accuracy. Moreover, we extract further depth cues from the motion between intra-capture gated frames that are sequentially acquired for different gates.

\vspace{-0.12cm}
\section{Gated Imaging} \label{sec:gated_imaging}
\begin{figure}[!t]
    \centering
    \vspace{-1eM}
    \includegraphics[width=0.49\textwidth]{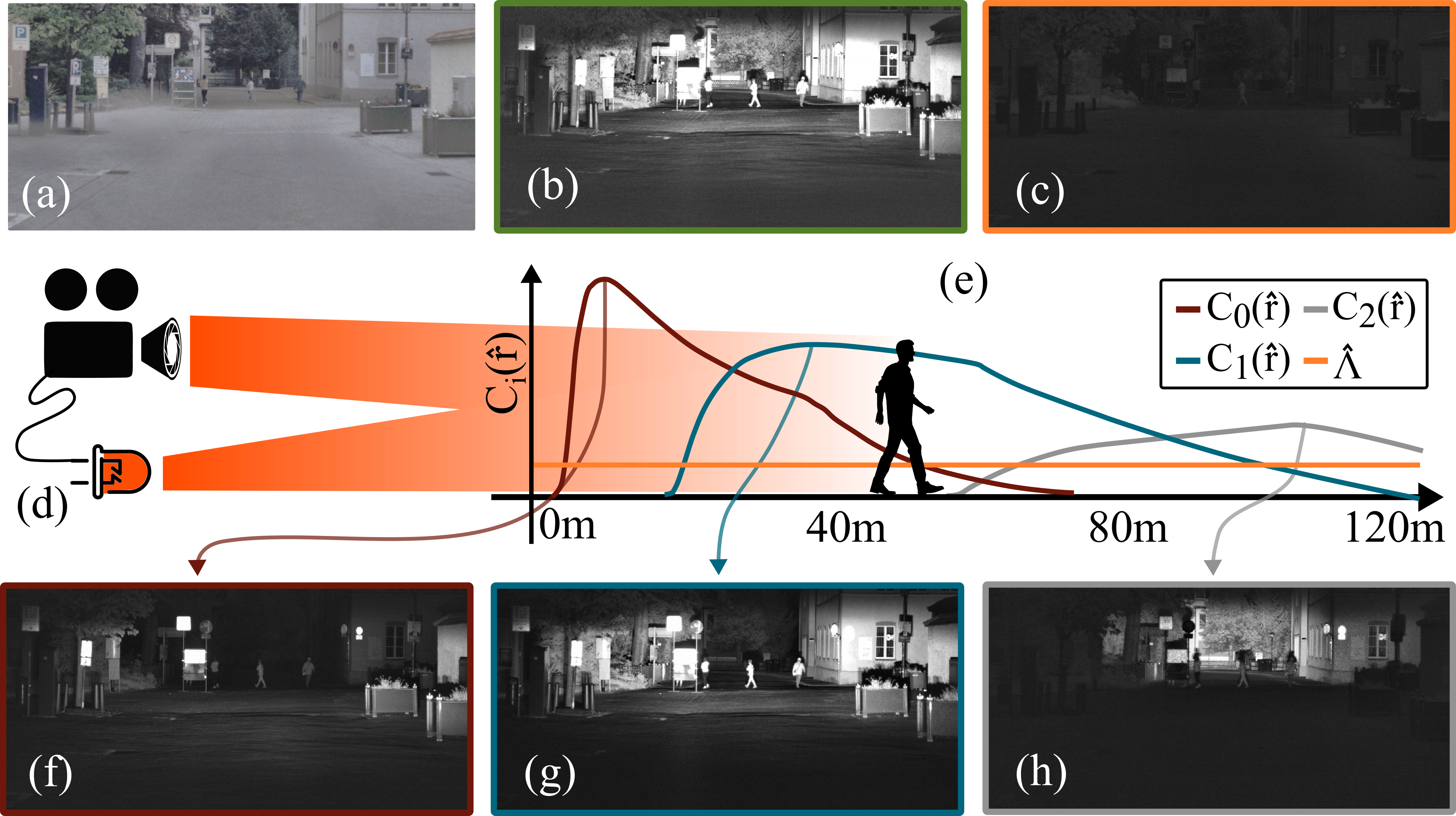}
    \vspace{-1.5eM}
    \caption{A gated camera consists of a synchronized gated camera and a flash pulsed illumination source (d). Using different exposure gates, the image formation can be described with three range-intensity profiles $C_{i}$, $i \in \{0,1,2\}$ plotted depending on distance $r[m]$, and an ambient, unmodulated $\lhat$ scene contribution. An overlay of all exposures is shown in \textcolor{dai_green}{green} (b). Individual range intensity profiles are shown for (f) short distance \unit[3-72]{m} in \textcolor{dai_deepred}{deepred}, (g) moderate distance \unit[18-123]{m} in \textcolor{dai_petrol}{petrol}, (h) far distance \unit[57-176]{m} in \textcolor{dai_gray}{gray} and (c) ambient light ($\Z_t^p$) in \textcolor{dai_orange}{orange}. A corresponding RGB capture of the scene is shown in (a).}\vspace{-1.eM}
    \label{fig:rip}
\end{figure}
\vspace{-0.25eM}
Before discussing our proposed method, we briefly review the principles of gated imaging. Figure~\ref{fig:rip} illustrates a gated imaging system which consists of a synchronized camera and flash illumination source.  In contrast to scanning LiDAR systems, the flash illumination source illuminates the scene 
through a laser pulse $p$, before capturing the reflected light echo with a $\xi$ delay. 
The reflected light is captured through a CMOS imaging sensor, which only captures 
photons arriving in a given temporal gate with profile $g$. Following 
Gruber \textit{et al.}~\cite{gruber2018learning}, we denote 
a single gated exposure as,
\begin{align}
Z_t^i(r)\;&=\;\alpha\,C_i(r)\\
    \;&=\;\alpha \int\limits_{-\infty}^{\infty} g_i(t-\xi)p_i\left(t\,-\,\cfrac{2r}{c}\right)\beta(r)dt,
\label{eq:gated_img}
\end{align}
where $Z_t^i(r)$ is the gated exposure, indexed by $i$, at distance $r$ and time $t$;
$C_i(r)$ is the range intensity profile, i.e., the convolution of 
the gated slice and its corresponding pulse profile; 
$\alpha$ is the surface reflectance (albedo), and $\beta$ the attenuation along a given path due to atmospheric 
interactions. The depth dependent path attenuation becomes 
unity in the absence of any participating media.

We note that during daytime, this model is incomplete due the high spectral solar power within the 
NIR band that leads to a significant number of unmodulated 
photons captured as an ambient light $\Lambda$ component. We modifiy the model from ~\cite{gruber2018learning} as
\begin{align}
Z_t^i(r)\;=\;\alpha\,C_{i}(r)+\,\Lambda.
\label{eq:final_gated_eq}
\end{align}
Similar to other CMOS-based sensing methods, gated imaging is also affected by noise, that can be modelled with a signal-dependent Poisson $\eta_{p}$ and Gaussian $\eta_{g}$, resulting in,
\begin{align}
Z_t^i\;=\;\alpha\,C_{i}(r)+\,\Lambda\,+\,\eta_{g}+\,\eta_{p}.
\label{eq:final_gated_eq}
\end{align}
In this work, at a given time $t$, we capture three sequential gated slices with spatial resolution $H \times W$ as $\mathbf{Z}^{i}_{t}\,\in \mathbb{R}_{+}^{H \times W}$ with delays $\xi_{\{0,1,2\}}$ and an unmodulated NIR passive image $\Z_t^p$ as illustrated in Figure~\ref{fig:rip}. With a native frame rate of 120~Hz, the proposed gated camera provides a full set of observations at 30~Hz.

\section{Self-Supervised Gated Depth Estimation}
\begin{figure*}[!t]
\vspace*{-2mm}
    \centering
    \vspace{-1eM}
    \includegraphics[width=0.99\textwidth]{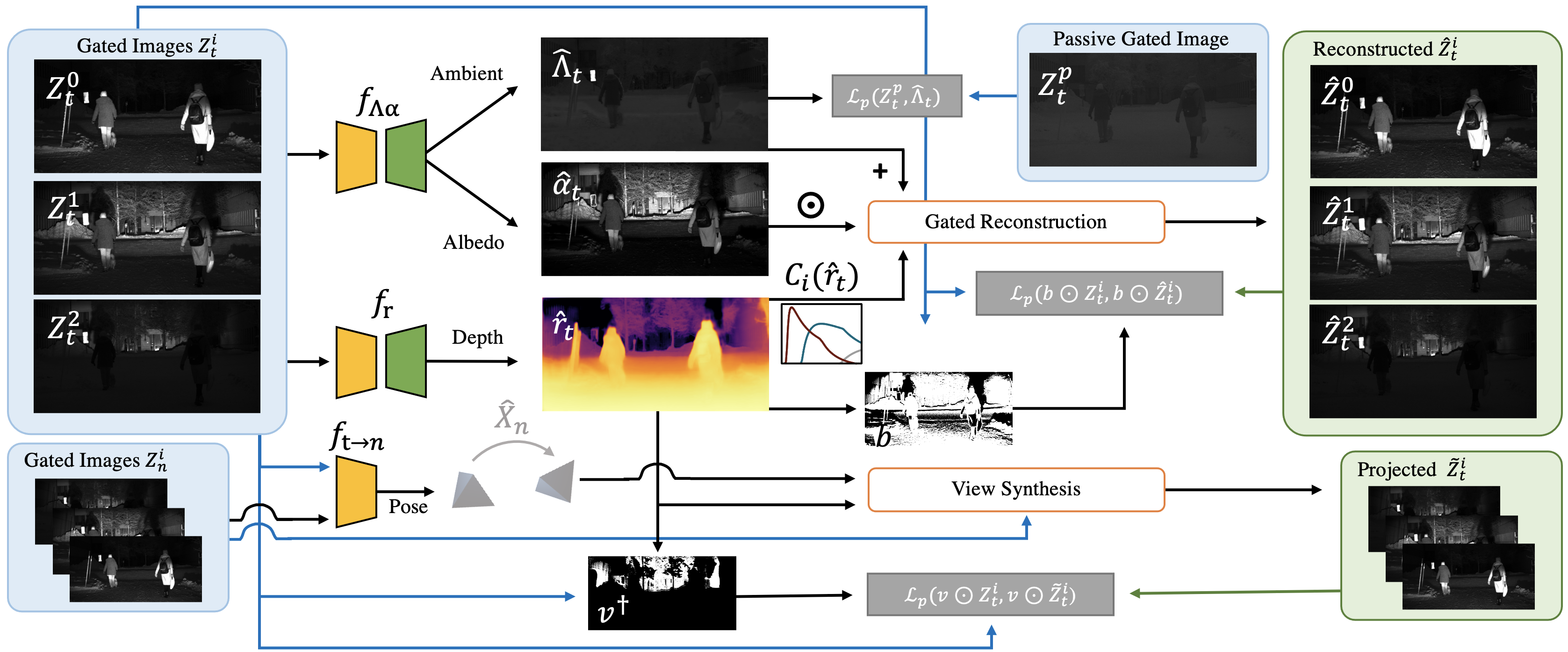}
    \vspace{-5pt}
    \caption{The proposed self-supervised Gated2Gated architecture estimates dense depth from a set of three gated images, by learning from cyclic gated and temporal consistency. %
    Thereby, the inherent scale ambiguity is solved through the range intensity profiles introducing a scale cue during training. Additional, mask $b,v$ resolve multi-path and shadow artifacts breaking Eq.~\eqref{eq:gated_reconstruction}.
    }
    \label{fig:Architecture}
    \vspace{-0.3cm}
\end{figure*}
The proposed method learns to predict depth $\rhat_t$ without ground 
truth supervision from LiDAR or simulation. To this end, we 
exploit the cyclic measurement consistency of gated images 
and temporal consistency in the depth predictions. 
Self-supervision allows us to overcome the limited depth 
range (\unit[80]{m}) of methods 
trained on LiDAR ground-truth and removes complex synchronization 
processes between LiDAR and cameras. Furthermore, we can train 
our models on harsh weather conditions, e.g., fog, rain, or snow,
where LiDAR-based ground-truth fails. 

Our proposed Gated2Gated architecture is 
illustrated in Figure~\ref{fig:Architecture}. 
While our model is general in terms of the input gated slices, we consider 
three slices $\Z_t^i$, for $i={0, 1, 2}$, at each time $t$. 
The gated measurements $\Z_t^i$
are concatenated in a tensor $\Z_t$ that is fed to three convolutional neural networks 
that disentangle the input into albedo, ambient light, and depth, 
which are then used to reconstruct the input slices
using a cyclic loss. In addition to this novel gated imaging-based 
training signal, we exploit temporal consistency between 
temporally adjacent gated frames to handle regions with shadows and multi-path reflections.

Specifically, the proposed architecture is composed of three networks.
The first network predicts a dense depth map per gated tensor 
$\Z_t$, denoted as $f_{\mathbf{r}}: \Z_t \to \rhat_{t}$.
The second network also takes $\Z_t$ as input, and predicts ambient 
and albedo, denoted as  %
$f_{\mathbf{\Lambda\alpha}}: \Z_t \to (\lhat_t, \ahat_t)$.
The third network takes two temporally adjacent gated tensors as 
input $(\Z_t, \Z_n)$, and 
predicts a rigid 6 DoF pose transformation $\xhat$ from $\Z_n$ to $\Z_t$, 
denoted as
$   \xhat_n=\begin{pmatrix}
    R_{\mathbf{t\rightarrow n}} & t_{\mathbf{t\rightarrow n}} \\
    0 & 1 
    \end{pmatrix}$,
with $R_{\mathbf{t\rightarrow n}} \in SO(3)$ and $t_{\mathbf{t\rightarrow n}} \in \mathbb{R}^{3\times1}$ generated by %
$f_{\mathbf{t\rightarrow n}}: \left(\Z_t, \Z_n\right) \to \xhat_{\mathbf{t\rightarrow n}}$.\vspace{0.1cm}

The learned function $f_{\mathbf{r}}$ is optimized to predict the absolute depth value and is supervised using the other two auxiliary functions, $f_{\mathbf{\Lambda\alpha}}$ and $f_{\mathbf{t\rightarrow n}}$. The first auxiliary function is used to exploit cyclic measurement consistency with the measured gated slice, i.e., enforce that the predicted depth is consistent with a gated measurement. The second auxiliary function allows us to exploit temporal consistency between nearby gated frames. Using these cues, the proposed method resolves scale ambiguity inherent in monocular depth estimation. The two consistency components are discussed in the following sections.

\begin{figure}[!t]
     \centering
     \vspace{-1eM}
     \begin{subfigure}{0.48\textwidth}
        \includegraphics[width=\textwidth]{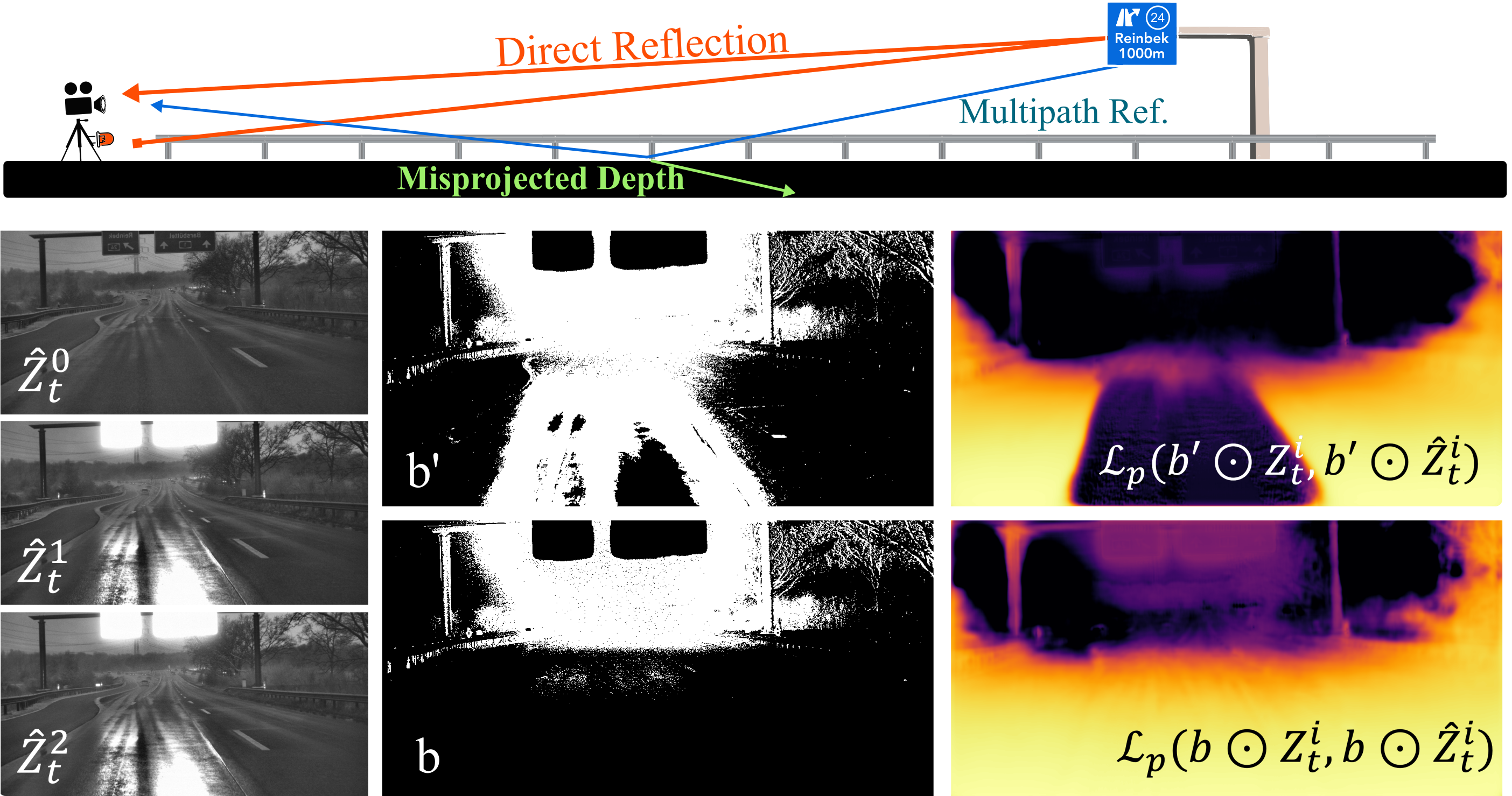} 
     \end{subfigure}
    \vspace{-0.5eM}
     \caption{\small Illustration of the valid pixel mask $b$. Retro reflective traffic signs reflect the illumination light back towards the camera (orange path), but spread out the light illuminating the ground causing multipath effects (blue path). This superposes the low intense groundplane intensity with the high intensity multi-path reflection containing further distant pulse information, leading to a wrong depth estimate (green path).
     Mitigation strategies are shown below using a valid pixel mask $b'$ (middle) and next to it the achieved depth prediction (right). The last row shows the refined mask $b$ (middle) and the corresponding training output (right).}\label{fig:MultiPathFiltering}
    \vspace{1eM} 
    \begin{subfigure}{0.48\textwidth}
        \includegraphics[width=\textwidth]{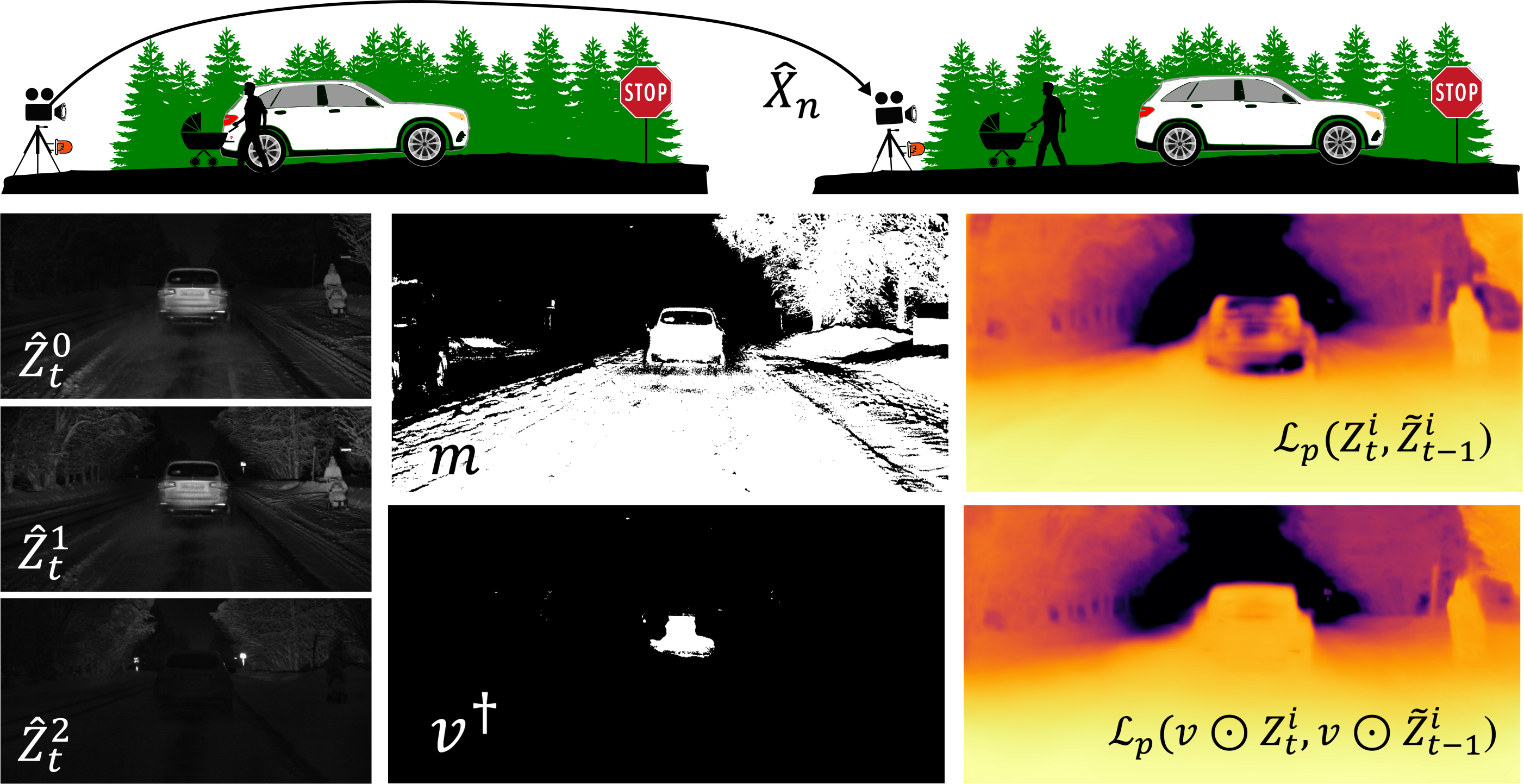}
    \end{subfigure}
    \vspace{-0.5eM}
    \caption{\small The top row shows two temporally adjacent frames and their rigid transformation $\hat{\mathbf{X}}_{n}$ as motivation for the valid mask $v$. Note that moving objects that remain in the same place in both frames are indistinguishable from objects at an infinite distance, causing holes in the predicted depth maps (middle right). Estimating close pixels in $m$ (middle) infinite distances can be filtered according to the mask $v$ (last row middle) with corresponding depth results (last row right).}\label{fig:ClosePixelFiltering}\vspace{-1eM}
\end{figure}

\vspace{-0.14cm}
\subsection{Cyclic Gated Consistency}\label{sec:GatedCyclcImageReconstruction}
\vspace{-0.3eM}The cyclic gated consistency loss supervises the predicted depth $\rhat_t$, ambient $\lhat_t$ and albedo $\ahat_t$, by reconstructing the gated slices. To this end, we use a simplified version of 
Equation~\ref{eq:final_gated_eq}, in which we model 
ambient and noise together, that is
\begin{equation}\label{eq:ambient_term}
    \lhat = \,\eta_{g}+\,\eta_{p}+\,\Lambda.
\end{equation}
The final model then can be written as
\begin{equation}\label{eq:gated_reconstruction}
    \Zhat_t^i = \ahat_t C_i(\rhat_t) + \lhat_t ,
\end{equation}
with $C_i$ being range-intensity profiles (as defined in Figure \ref{fig:rip}). 
The $C_i$ profiles are measured experimentally with 
calibrated targets and approximated with Chebyshev polynomials $T_n$
\begin{equation}
    T_0=1,\quad T_1=x, \quad T_{n+1}=2xT_{n}-T_{n-1},
\end{equation}
up to order of $N=6$.
The ambient predictions $\lhat$ can be directly supervised using the ground truth captured passive images $\Z_t^p$ and the photometric loss $\mathcal{L}_p$.   %

As loss function $\mathcal{L}_p$ \cite{godard2017unsupervised}, we use Structural Similarity (SSIM) \cite{ssim} and $\mathcal{L}_1$ norm:
\begin{equation}\label{eq:gated_reconstruction_loss}
    \mathcal{L}_p(\Z_t^i, \Zhat_t^i) = 0.85\cdot\frac{\scriptstyle 1-SSIM(\Z_t^i, \Zhat_t^i)}{ \scriptstyle 2} + 0.15 \cdot ||\scriptstyle{\Z_t^i-\Zhat_t^i}||_1.
\end{equation}
\vspace{-2.2eM}
\paragraph{Gated2Gated Cyclic Loss Masks.}
The gated slices $\Zhat_t^i$ can be reconstructed using the
proposed cyclic gated consistency, which enforces a match between a predicted depth estimate and the ground truth $\Z^i_t$ measurement it came from. Specifically, having predicted depth $r$ using $f_{\mathbf{r}}$ and $\alpha$ using $f_{\mathbf{\Lambda\alpha}}$, we can predict a gated image using Eq.~\eqref{eq:gated_reconstruction}. However, adopting a photometric loss~\cite{godard2017unsupervised,Zhou2017} between the measurement and the prediction fails in practice as severe multipath effects, missing illumination due to occlusion, and saturation due to retro-reflective signs can break the model in Eq.~\eqref{eq:gated_reconstruction}. We illustrate 
these issues in Figure~\ref{fig:MultiPathFiltering}. To this end, we introduce the following pixel masks that optimize 
the performance of the cyclic self-supervised model in those conditions.

\textit{Pixel Variance.}
Pixels $p_{xy}$ with similar intensity across all slices are not 
modulated and are either beyond the range of the illumination, 
or highly absorptive. We then define a mask 
to filter out pixels with low variance as follows,
\begin{equation}
    D_{xy} \coloneqq \{\,(x,y) \mid \left(\max_{i}\left(p_{xy}^i\right)-\min_{i}\left(p_{xy}^i\right)\right) > \theta \,\}.
\end{equation}

\textit{Saturated Pixels.}
We also exclude saturated pixels, i.e., pixels with 
high-intensity values in all three gated slices,
using the following mask,

\begin{equation}
    M_{xy} \coloneqq \{\,(x,y) \mid \max_{i}(p_{xy}^i)< \gamma \,\}.
\end{equation}
The pixel variance and saturated pixels masks are combined into
a binary mask $b_{xy}'$, defined as,
\begin{align}
    b_{xy}'=\begin{cases} 
                        1 & \mathrm{if}\ (x,y) \in D_{xy} \wedge (x,y) \in M_{xy}  \\
                       0 & \mathrm{otherwise} .
                       \end{cases}
\end{align}

\textit{Multipath Correction.}
The binary mask $b'$ is further refined by modeling 
multipath effects, taking advantage of the view geometry, 
as illustrated in Figure~\ref{fig:MultiPathFiltering}. 
In automotive scenes, the most severe multipath effects result 
from reflective road surfaces. Using the camera intrinsics, 
we estimate a conservative constant ground plane with 
normal $n$ and height $h$. Furthermore, we estimate an approximated depth measurement $\mathbf{\Tilde{r}}$ by comparing the intensity values of the three gated slices. This allows us to filter out 
pixels $(x,y)$ that get back-projected to 3D coordinates 
substantially lower than the ground plane:
\begin{equation}
    E_{xy} \coloneqq \{\,(x,y) \mid (\mathbf{\rhat} K^{-1}x_t)n<h \,\},
\end{equation}
where $x_t=[x,y,1]$ denote homogeneous pixel coordinates and $K$
denote the camera matrix. 
The final Gated2Gated cyclic loss mask $b$ is then defined as, 
\begin{align}\vspace{-0.1eM}
    b_{xy}=\begin{cases} 
                        1 & \mathrm{if}\ (x,y) \notin E_{xy} \wedge b_{xy}'=1  \\
                       0 & \mathrm{otherwise} ,
                       \end{cases}
\end{align}\vspace{-0.1eM}
and the cyclic loss function is defined as,
\begin{equation}
    \mathcal{L}_{cyc} = \sum_{i=0}^2\mathcal{L}_p(b\odot\Z_t^i, b\odot\Zhat_t^i) + \mathcal{L}_p(\lhat,\Z_t^p).
\end{equation}

\vspace{-0.14cm}
\begin{figure*}[h!]
\vspace*{-8mm}
	\centering
	\begin{minipage}[t]{0.49\linewidth}
		\scriptsize
		\subcaption{\scriptsize Clear Day: Gated2Gated mitigates artefacts in depth map at far distances.  }
		\vspace{0.05cm}
		\renewcommand{\arraystretch}{0.7}
\setlength{\tabcolsep}{2pt}
\begin{tabular}{@{}ccc@{}}
	RGB & 
	Full Gated &
	LiDAR \\
	
	\includegraphics[width=0.31\columnwidth]{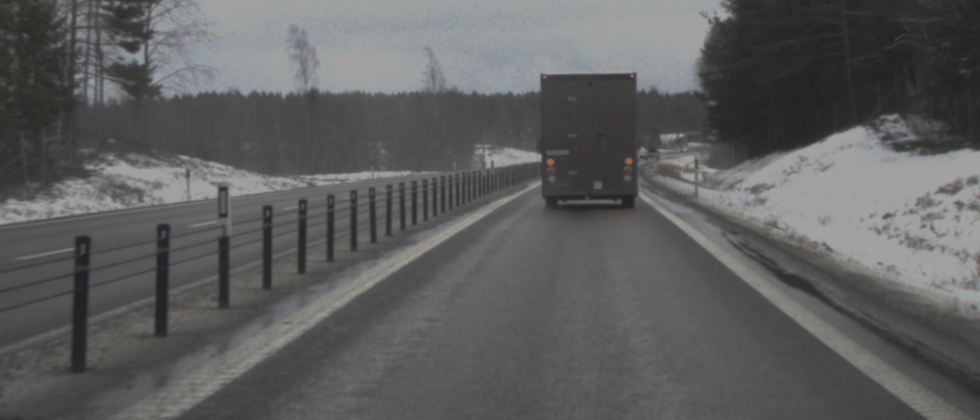} &
	\includegraphics[width=0.31\columnwidth]{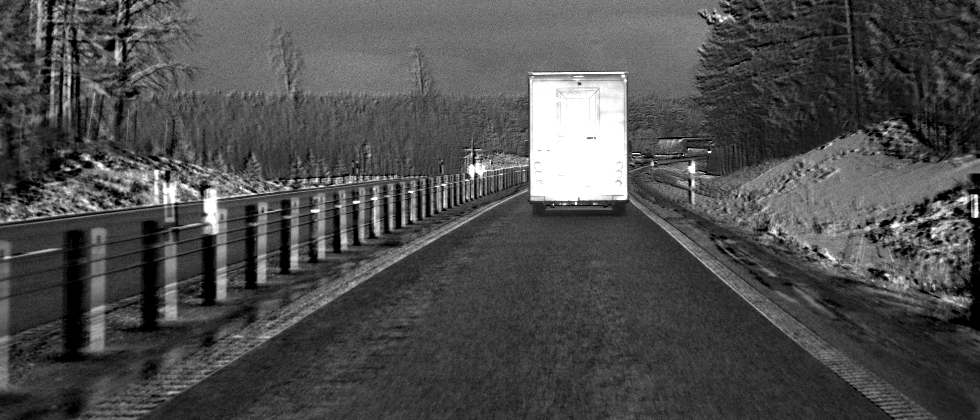} &
	\includegraphics[width=0.31\columnwidth]{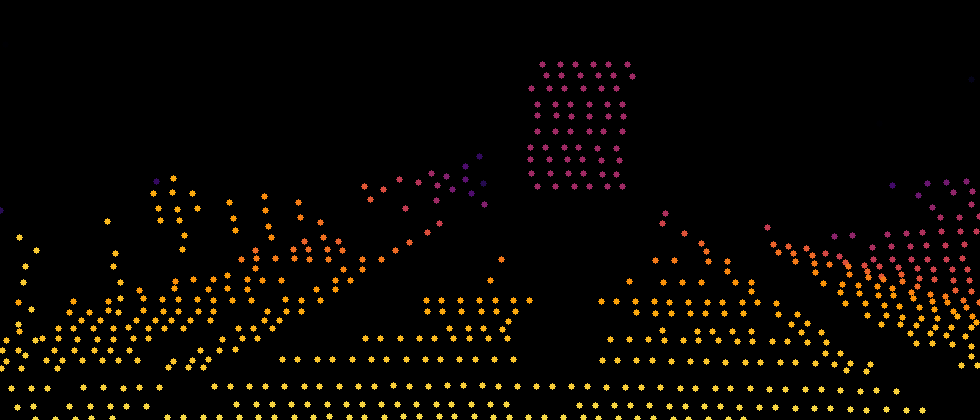} \\

    \textbf{Gated2Gated} &
	Gated2Depth\cite{gated2depth2019} &
    LiDAR+RGB  \cite{ma2018sparse} 
    \\
    
	\includegraphics[width=0.31\columnwidth]{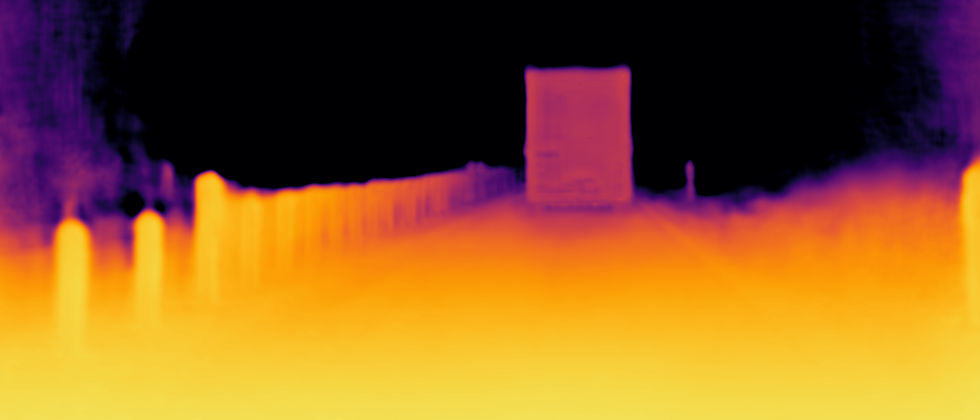} &
	\includegraphics[width=0.31\columnwidth]{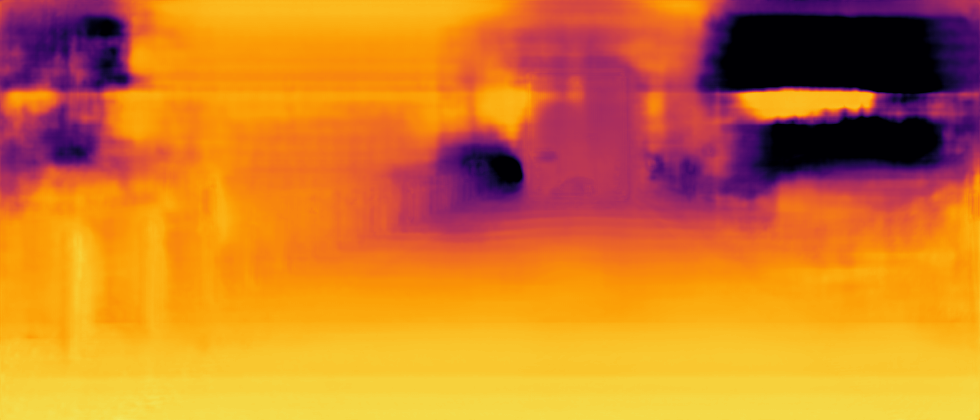} &
	\includegraphics[width=0.31\columnwidth]{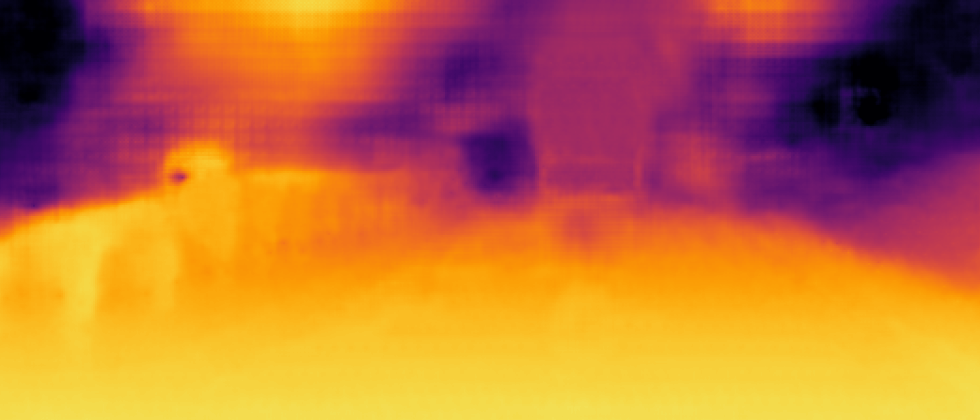} 
    \\
    
	Stereo-SGM \cite{Hirschmuller2008} &
	PSM-Net \cite{Chang2018} & 
    PackNet-Sfm \cite{guizilini20203d} 
	\\
	
	\includegraphics[width=0.31\columnwidth]{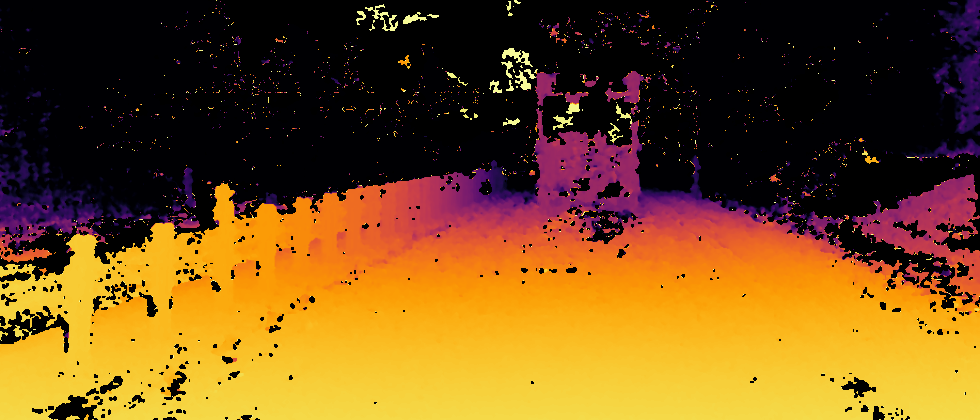} &
	\includegraphics[width=0.31\columnwidth]{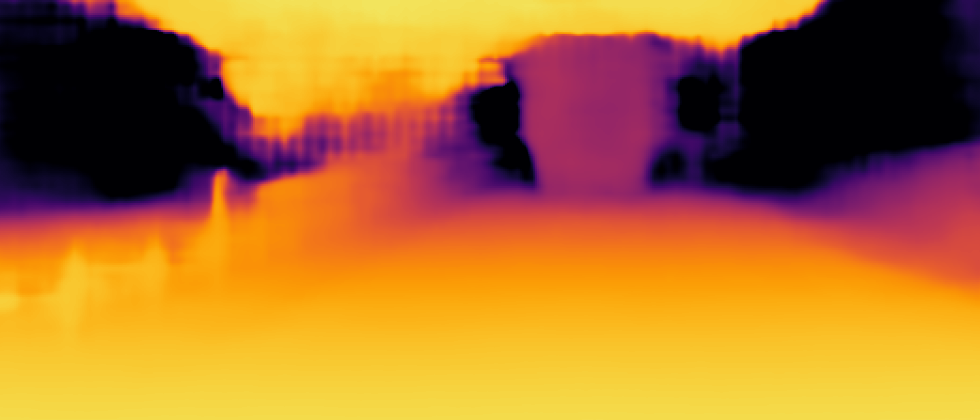} &
	\includegraphics[width=0.31\columnwidth]{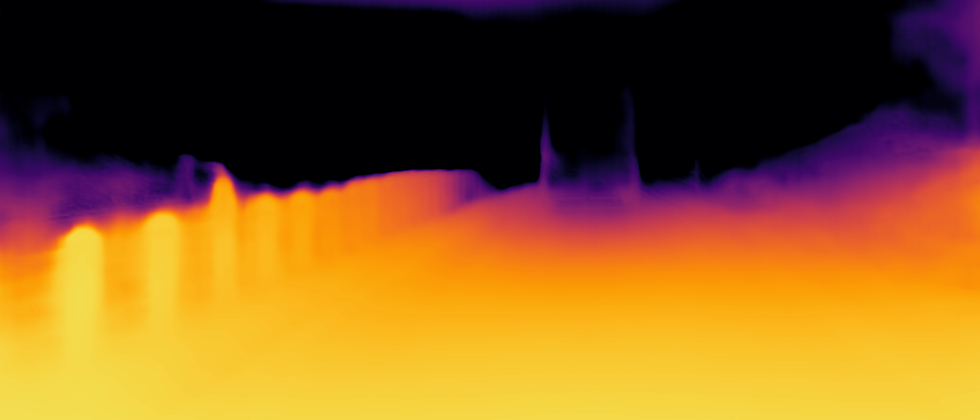} 
	\\
\end{tabular}

		\label{fig:comp_ref_methods_day}
	\end{minipage}%
	\centering
	\begin{minipage}[t]{0.49\linewidth}
		\scriptsize
		\subcaption{\scriptsize Snow: Gated2Gated provides more robust depth predictions in adverse weather.}
		\vspace{0.05cm}
		\renewcommand{\arraystretch}{0.7}
\setlength{\tabcolsep}{2pt}
\begin{tabular}{ccc}
	RGB & 
	Full Gated &
	LiDAR \\
	
	\includegraphics[width=0.31\columnwidth]{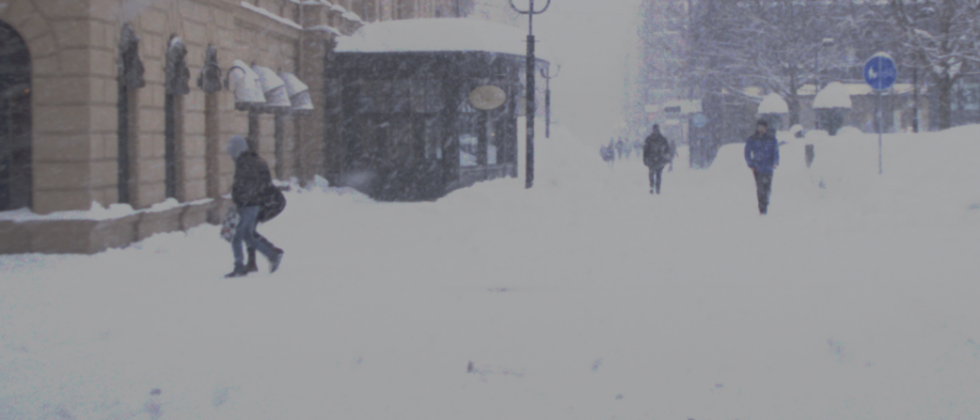} &
	\includegraphics[width=0.31\columnwidth]{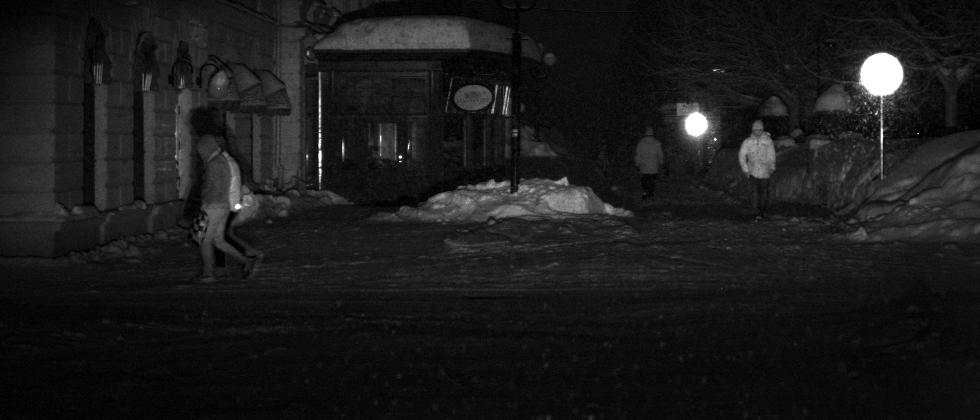} &
	\includegraphics[width=0.31\columnwidth]{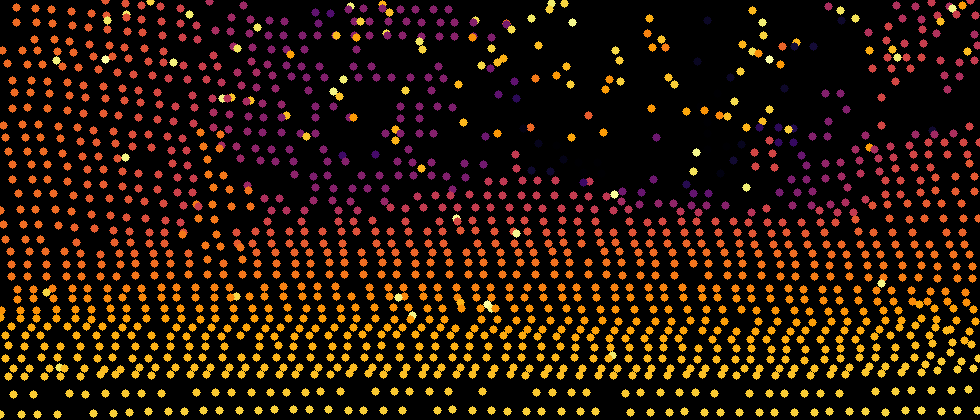} 
	\\
	
	\textbf{Gated2Gated} &
	Gated2Depth\cite{gated2depth2019} &
	LiDAR+RGB  \cite{ma2018sparse} \\

	\includegraphics[width=0.31\columnwidth]{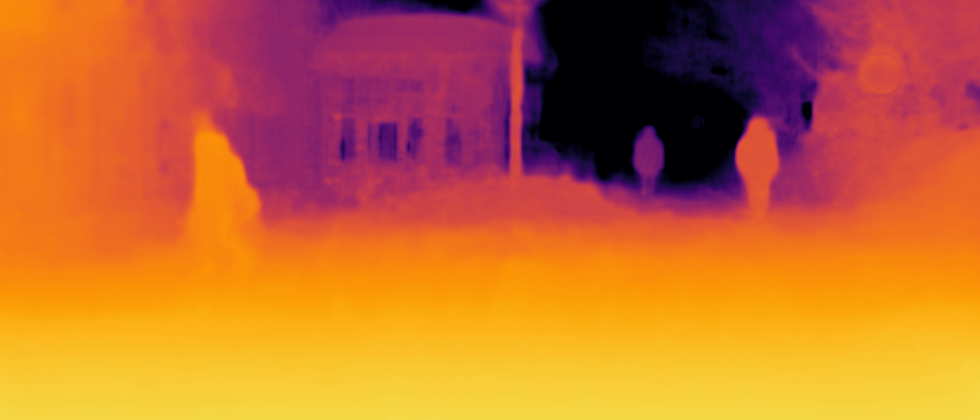} &
	\includegraphics[width=0.31\columnwidth]{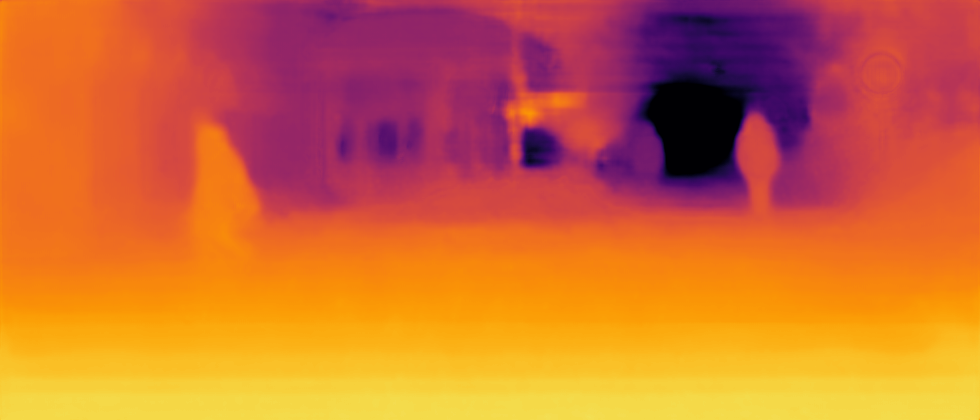} &
	\includegraphics[width=0.31\columnwidth]{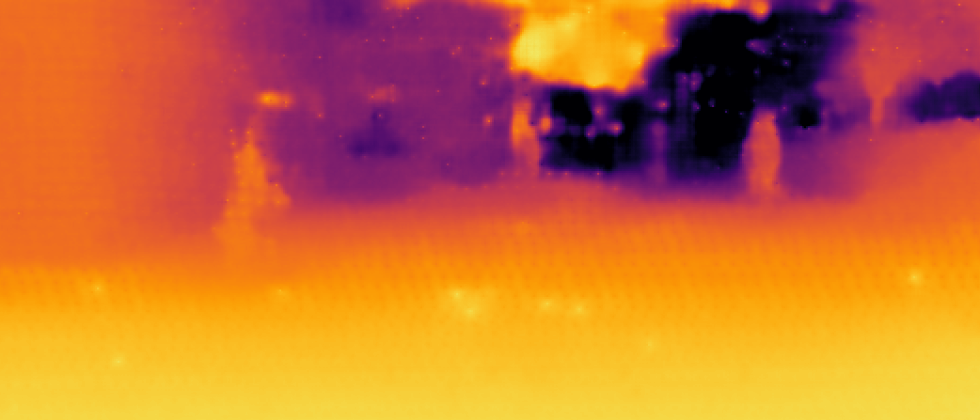} 
	\\

	Stereo-SGM \cite{Hirschmuller2008} &
	PSM-Net \cite{Chang2018} & 
    PackNet-Sfm \cite{guizilini20203d} 
	\\
	
	\includegraphics[width=0.31\columnwidth]{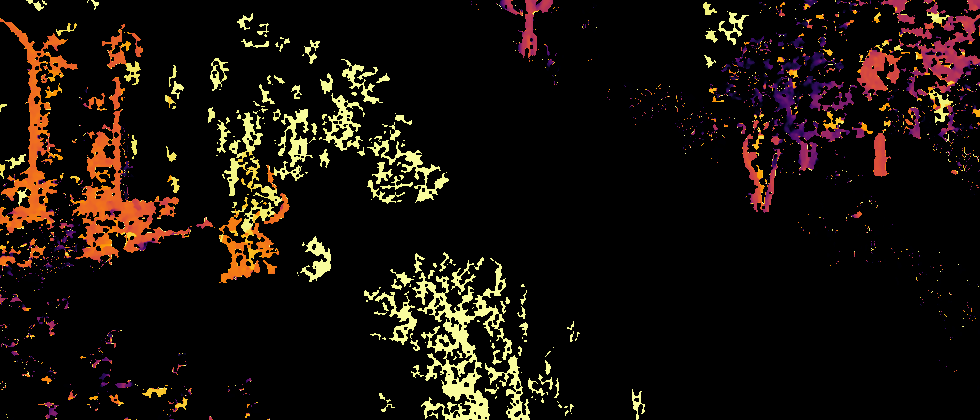} &
	\includegraphics[width=0.31\columnwidth]{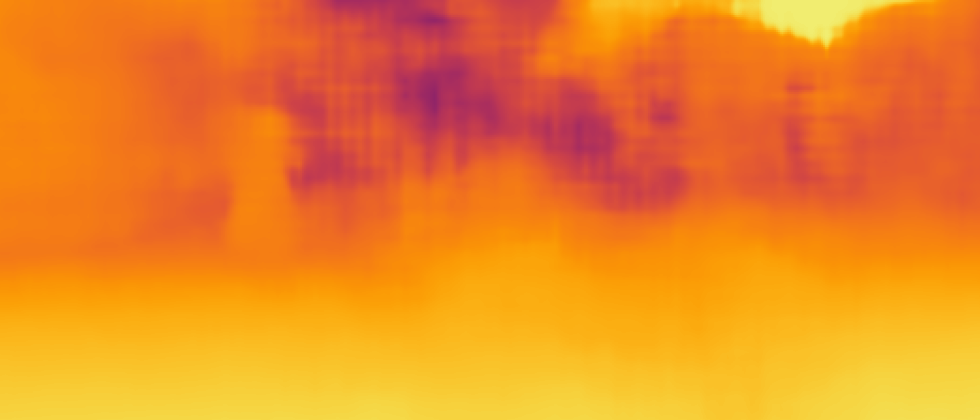} &
	\includegraphics[width=0.31\columnwidth]{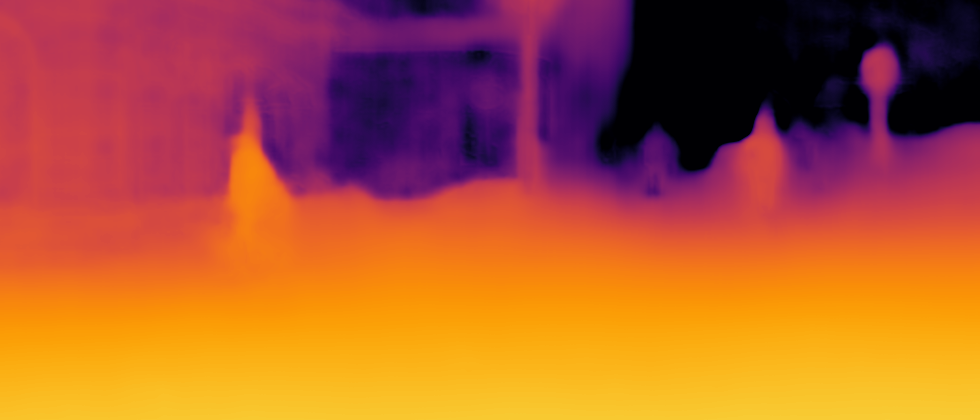} 
	\\
\end{tabular}

		\label{fig:comp_ref_methods_snow}
	\end{minipage}%
	\centering
	\begin{minipage}[t]{0.02\linewidth}
		\scriptsize
		\begin{tabular}{@{}>{\centering\arraybackslash}m{0.2cm}}
			\multirow{1}{*}[-0.74cm]{\hspace{-0.042cm}[m]} \\
			\multirow{3}{*}[-0.72cm]{\hspace{-0.2cm}\includegraphics[height=3.95cm]{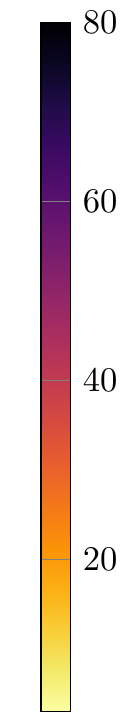}}\\
			
		\end{tabular}
	\end{minipage}%
	\vspace*{-2mm}
	\caption{\textbf{Qualitative comparison of Gated2Gated and existing methods}. For two examples (a) clear day and (b) snow day, Gated2Gated predicts sharper depth maps than existing methods. (Full-Gated image refers here to an integral image $\sum_{i=0}^{2} \Z_t^i-2\Z_t^p$).}
	\label{fig:comp_ref_methods}
	\vspace{-0.4cm}
\end{figure*}
\subsection{Temporal Depth Consistency}\label{sec:GatedTimeSupervision}

As illustrated in Figure~\ref{fig:Architecture}, we use 
view synthesis to introduce temporal consistency 
between adjacent gated images during training. 
Specifically, we reconstruct the view of central gated image $\Z_t$ from temporal neighbors $\Z_n$ using the camera matrix $K$, the predicted depth $\rhat_t$ and camera pose transformation $\hat{\mathbf{X}}_{t\rightarrow n}$. Considering $x_t$ and $x_n$ homogeneous
(pixel) coordinates from $\Z_t^i$ and $\Z_n^i$, 
the mapping from source pixels $x_t$ to target pixels 
$x_n$ is defined as follows 
\begin{equation}
    x_{n} \sim K\hat{\mathbf{X}}_{t\rightarrow n} \rhat_t K^{-1}x_t.   
\end{equation}
Similar to \cite{Zhou2017}, we compare the reconstructed view $\hat{\Z}_{t}$ with $\Z_t$ using photometric loss and use it to train the depth prediction network $f_{\mathbf{r}}$.  Unfortunately, this naive approach fails in the presence of moving occlusions due to ego-motion and the movement of non-stationary objects, which violate the rigid pose transformation. As in the case 
of the cyclic loss, we introduce validity masks as illustrated in Figure~\ref{fig:ClosePixelFiltering}.

\vspace{-1eM}
\paragraph{Infinity Correction Masks.}
To handle the dynamic scene objects, we use gated 
illumination cues to prevent projections to infinity 
(see Figure~\ref{fig:ClosePixelFiltering}). 
Specifically, we define two valid sets per pixel 
position $(x,y)$. 
The first set $S^1$ analyzes the pixel relation between first $\Z_t^0$ and last slice $\Z_t^2$. This allows to find valid pixels in a range from \unit[3-$s_0$]{m}, where $\C_0(s_0)=c\C_2(s_0)$. The second set $S^2$ analyses the distance in the medium ranges \unit[18-$s_1$]{m} (18~m due to the specific gate selection we made, see Supplemental Document) comparing the second $\Z_t^1$ and last slice $\Z_t^2$  with $\C_1(s_1)=c\C_2(s_1)$. In terms of measured pixel intensities the sets can be given as,
 \begin{align}
     S_{xy}^1 \coloneqq& \{\,(x,y) \mid \Z_t^0(x,y) \geq c\cdot \Z_t^2(x,y) \,\},\\
     S_{xy}^2 \coloneqq& \{\,(x,y) \mid \Z_t^1(x,y) \geq c\cdot \Z_t^2(x,y) \,\}.
 \end{align}
This allows us to filter points illuminated by the last slice ranging from \unit[$\max(s_0,s_1)$-176]{m} (specific to our gates again) and to find close regions with a binary mask $m$,
\begin{align}
    m_{xy}\,=\, \begin{cases} 
                        1 & \mathrm{if}\ (x,y) \in S_{xy}^1\ \lor (x,y) \in S_{xy}^2  \\
                       0 & \mathrm{otherwise} .
                       \end{cases}
 \end{align}
 Then for the pixels with $P_{xy} \coloneqq \{\,(x,y) \mid m_{xy}=1\,\}$ the average median $\bar{\mathbf{r}}_t$ is calculated. Comparing the median $\bar{\mathbf{r}}_t$ to the predicted depth $\rhat_t$ allows us to filter mispredicted depth values stemming from moving objects which cannot be captured by the rigid transformation $\xhat_n$, see Figure~\ref{fig:ClosePixelFiltering}. Specifically, we omit all depth values twice the average median leading to a binary mask $v_{xy}$,
 \begin{align}
\centering
         v_{xy}\,= \begin{cases} 
                        1 & \mathrm{if}\ (x,y) \in V_{xy}  \\
                       0 & \mathrm{otherwise} ,
                     \end{cases}
\end{align}
with set $V_{xy} \coloneqq \{\,(x,y) \mid m_{xy}=1\ \wedge\  \rhat_t(x,y)<2\cdot\bar{\mathbf{r}} \,\}$. %

Additionally, scene points that are only visible either in the source or the target image break our image formation model.
Such scenarios can be caused by foreground occluders obstructing the view of the background. 
In dynamic scenes, this occlusion changes in each timestep. For a triplet of adjacent frames $(t-1, t, t+1)$, we can define the minimum pixel error out of those pairwise differences, as occlusions cause a higher re-projection error. This can be explained as occluder and background have higher textural differences than neighboring pixels in the background. 
Therefore, we calculate minimum of per-pixel loss between the re-projection from two temporal adjacent pairs similar to \cite{godard2019digging} as,
\vspace{-4mm}
\begin{equation}
    \mathcal{L}_{temp}\,=\min_{n=\lbrace t-1,t+1\rbrace}\sum_{i=0}^2\left(\mathcal{L}_p\left(v\odot\Z_t^i, v\odot\Zhat_{t}^i(n)\right)\right).
\end{equation}
\vspace{-1eM}

\begin{table}[!t]
    \footnotesize
    \setlength{\tabcolsep}{4pt} %
    \setlength\extrarowheight{2pt}
    \centering
    \resizebox{.99\linewidth}{!}{
    \begin{tabular}{@{}c|lccccccccc@{}}
            \toprule
            & \multirow{2}{*}{\textbf{\textsc{Method}}} & \textbf{Modality} & \textbf{Train} & \textbf{RMSE}     & \textbf{ARD}   & \textbf{MAE}  & $\boldsymbol{\delta_1}$ & $\boldsymbol{\delta_2}$ & $\boldsymbol{\delta_3}$ & \textbf{Compl.} \\ 
			&& &  & $\left[ m \right]$  &  & $\left[ m \right]$ & $\left[ \% \right]$ & $\left[ \% \right]$ & $\left[ \% \right]$ & $\left[ \% \right]$\\
			\midrule
			\multicolumn{11}{c}{\textbf{Real Data -- Night (Evaluated on Lidar Ground Truth Points)}} \\
			\midrule
			\multirow{7}{*}{\rotatebox[origin=l]{90}{\parbox[c]{2.8cm}{\centering \textbf{\textsc{Supervised}}}}}   & \textsc{PSMNet \cite{Chang2018}} & Stereo-RGB & D & 14.58 & \underline{0.21} & 8.34 & 68.75 & 82.63 & 89.36 & 100 \\ 
  & \textsc{SGM \cite{Hirschmuller2008}} & Stereo-RGB & - & 15.51 & 0.36 & 8.75 & 63.94 & 76.19 & 82.31 & 63 \\ 
  & \textsc{Sparse-to-Dense \cite{ma2018sparse}} & Lidar(GT)+RGB & D & \underline{8.79} & \underline{0.21} & \underline{4.38} & \textbf{87.64} & \textbf{93.74} & \textbf{95.88} & 100 \\ 
  & \textsc{Regression Tree} \cite{adam2017bayesian} & Gated & D & 10.54 & 0.24 & 6.01 & 76.73 & 89.74 & 93.45 & 40 \\ 
  & \textsc{Least Squares} & Gated & - & 13.13 & 0.42 & 8.88 & 43.60 & 55.80 & 63.54 & 31 \\ 
  & \textsc{Gated2Depth \cite{gated2depth2019}} & Full-Gated & D & 14.86 & 0.29 & 8.84 & 58.79 & 58.79 & 79.84 & 100 \\ 
  & \textsc{Gated2Depth \cite{gated2depth2019}} & Gated & D & \textbf{8.39} & \textbf{0.15} & \textbf{3.79} & \underline{87.52} & \underline{93.00} & \underline{95.21} & 100 \\

			\midrule
			\multirow{7}{*}{\rotatebox[origin=l]{90}{\parbox[c]{2.8cm}{\centering \textbf{\textsc{Unsupervised}}}}}   & \textsc{Monodepth \cite{Godard2017}} & RGB & S & 11.41 & 0.23 & 6.18 & 76.64 & \underline{89.53} & 94.19 & 100 \\ 
  & \textsc{Monodepth \cite{Godard2017}} & Full-Gated & S & 15.41 & 0.52 & 11.33 & 31.72 & 71.23 & 88.74 & 100 \\ 
  & *\textsc{Packnet} \cite{guizilini20203d} & RGB & M & 12.15 & 0.27 & 6.87 & 69.14 & 86.93 & 92.57 & 100 \\ 
  & *\textsc{Packnet-Slim} \cite{guizilini20203d} & Gated & M & \underline{10.78} & \underline{0.22} & 6.02 & 74.37 & 89.44 & \underline{94.34} & 100 \\ 
  & *\textsc{Monodepth2} \cite{godard2019digging} & RGB & M & 14.92 & 0.38 & 9.98 & 39.85 & 68.57 & 83.99 & 100 \\ 
  & *\textsc{Monodepth2} \cite{godard2019digging} & Gated & M & 11.18 & 0.25 & \underline{5.99} & \underline{76.79} & 87.04 & 91.58 & 100 \\ 
  & \textbf{\textsc{Gated2Gated}} & Gated & MG & \textbf{9.43} & \textbf{0.21} & \textbf{4.86} & \textbf{82.17} & \textbf{91.54} & \textbf{94.48} & 100 \\

			\midrule
			\multirow{4}{*}{\rotatebox[origin=l]{90}{\parbox[c]{1.6cm}{\centering \textbf{\textsc{Ablation}}}}}  & \textbf{\textsc{Gated2Gated}} [$v$ \ding{55}, b \ding{55}] & Gated & MG & 10.05  &  0.27  &  5.36  &  80.06  &  90.44  &  93.75  &  100  \\ 
 & \textbf{\textsc{Gated2Gated}} [$v$ \ding{51}, b \ding{55}] & Gated & MG & 9.88  &  0.26  &  5.45  &  78.87  &  90.71  &  94.01  &  100  \\ 
 & \textbf{\textsc{Gated2Gated}} [$v$ \ding{55}, b \ding{51}] & Gated & MG & \underline{9.58 } & \underline{ 0.25 } & \underline{ 5.03 } & \underline{ 80.68 } & \underline{ 91.25 } & \underline{ 94.40 } &  100  \\ 
 & \textbf{\textsc{Gated2Gated}} [$v$ \ding{51}, b \ding{51}] & Gated & MG & \textbf{9.43 } & \textbf{ 0.21 } & \textbf{ 4.86 } & \textbf{ 82.17 } & \textbf{ 91.54 } & \textbf{ 94.48 } &  100  \\ 

			\midrule
			\multicolumn{11}{c}{\textbf{Real Data -- Day (Evaluated on Lidar Ground Truth Points)}} \\
			\midrule
			\multirow{7}{*}{\rotatebox[origin=l]{90}{\parbox[c]{2.8cm}{\centering \textbf{\textsc{Supervised}}}}}
			  & \textsc{PSMNet \cite{Chang2018}} & Stereo-RGB & D & 13.94 & 0.19 & 7.78 & 71.32 & 84.67 & 91.38 & 100 \\ 
  & \textsc{SGM \cite{Hirschmuller2008}} & Stereo-RGB & - & 9.63 & 0.17 & 4.59 & 85.80 & 92.72 & 95.20 & 86 \\ 
  & \textsc{Sparse-to-Dense \cite{ma2018sparse}} & Lidar(GT)+RGB & D & \underline{8.21} & \underline{0.16} & \underline{4.05} & \textbf{88.52} & \textbf{94.71} & \textbf{96.87} & 100 \\ 
  & \textsc{Regression Tree \cite{adam2017bayesian}} & Gated & D & 15.83 & 0.49 & 11.40 & 56.30 & 75.54 & 82.45 & 23 \\ 
  & \textsc{Least Squares} & Gated & - & 19.52 & 0.75 & 14.05 & 43.42 & 54.63 & 63.76 & 16 \\ 
  & \textsc{Gated2Depth \cite{gated2depth2019}} & Full-Gated & D & 13.75 & 0.26 & 8.16 & 62.48 & 62.48 & 82.93 & 100 \\ 
  & \textsc{Gated2Depth \cite{gated2depth2019}} & Gated & D & \textbf{7.61} & \textbf{0.12} & \textbf{3.53} & \underline{88.07} & \underline{94.32} & \underline{96.60} & 100 \\ 
			\midrule
			\multirow{7}{*}{\rotatebox[origin=l]{90}{\parbox[c]{2.8cm}{\centering \textbf{\textsc{Unsupervised}}}}}
			  & \textsc{Monodepth \cite{Godard2017}} & RGB & S & 10.24 & \underline{0.18} & 5.47 & 80.49 & 91.78 & 95.61 & 100 \\ 
  & \textsc{Monodepth \cite{Godard2017}} & Full Gated & S & 13.33 & 0.40 & 9.51 & 36.64 & 81.63 & 92.86 & 100 \\ 
  & \textsc{*Packnet \cite{guizilini20203d}} & RGB & M & 12.44 & 0.27 & 7.23 & 66.32 & 85.85 & 92.40 & 100 \\ 
  & \textsc{*Packnet-Slim \cite{guizilini20203d}} & Gated & M & 9.93 & \underline{0.18} & 5.34 & 78.98 & \underline{91.83} & \underline{96.06} & 100 \\ 
  & \textsc{*Monodepth2 \cite{godard2019digging}} & RGB & M & 13.18 & 0.33 & 8.79 & 44.99 & 75.87 & 90.65 & 100 \\ 
  & \textsc{*Monodepth2 \cite{godard2019digging}} & Gated & M & \underline{9.57} & \underline{0.18} & \underline{4.76} & \underline{83.20} & 91.75 & 94.94 & 100 \\ 
  & \textbf{\textsc{Gated2Gated}} & Gated & MG & \textbf{8.46} & \textbf{0.17} & \textbf{4.37} & \textbf{83.56} & \textbf{93.12} & \textbf{96.09} & 100 \\

			\midrule
			\multirow{4}{*}{\rotatebox[origin=l]{90}{\parbox[c]{1.6cm}{\centering \textbf{\textsc{Ablation}}}}}  & \textbf{\textsc{Gated2Gated}} [$v$ \ding{55}, b \ding{55}] & Gated & MG & 9.29  &  0.22  &  4.99  &  80.74  &  91.88  &  95.40  &  100  \\ 
 & \textbf{\textsc{Gated2Gated}} [$v$ \ding{51}, b \ding{55}] & Gated & MG & 9.14  & \underline{ 0.20 } &  4.99  &  80.82  &  92.26  &  95.58  &  100  \\ 
 & \textbf{\textsc{Gated2Gated}} [$v$ \ding{55}, b \ding{51}] & Gated & MG & \underline{8.85 } & \underline{ 0.20 } & \underline{ 4.75 } & \underline{ 81.20 } & \underline{ 92.57 } & \underline{ 95.85 } &  100  \\ 
 & \textbf{\textsc{Gated2Gated}} [$v$ \ding{51}, b \ding{51}] & Gated & MG & \textbf{8.46 } & \textbf{ 0.17 } & \textbf{ 4.37 } & \textbf{ 83.56 } & \textbf{ 93.12 } & \textbf{ 96.09 } &  100  \\ 

            \bottomrule
    \end{tabular}
    }
    \vspace*{-5pt}
    \caption{\label{tab:results_g2d}\small Comparison of our proposed framework and state-of-the-art methods on the Gated2Depth test dataset. We compare our model to supervised and unsupervised approaches. M refers to methods that use temporal data for training, S for stereo supervision, G for gated consistency  and D for depth supervision. *~marked method are scaled with LiDAR ground truth. Best results in each
    category are in \textbf{bold} and second best are
    \underline{underlined}.
	\vspace*{-4mm}
}
\end{table}
\begin{table*}[!t]
\vspace*{-4mm}
	\footnotesize
	\setlength{\tabcolsep}{2pt} %
	\centering
	\resizebox{1.0\linewidth}{!}{
		\begin{tabular}{l|lcccccccccccccccccccc}
			\toprule
			&& \multicolumn{5}{|c|}{\textbf{clear}} & \multicolumn{5}{|c|}{\textbf{light fog}} & \multicolumn{5}{|c|}{\textbf{dense fog}} & \multicolumn{5}{|c|}{\textbf{snow}}  \\
			&\textbf{\textsc{Method}}  & 
			RMSE & MAE & $\delta_1$ & $\delta_2$ & $\delta_3$ &
			RMSE & MAE & $\delta_1$ & $\delta_2$ & $\delta_3$ &
			RMSE & MAE & $\delta_1$ & $\delta_2$ & $\delta_3$ &
			RMSE & MAE & $\delta_1$ & $\delta_2$ & $\delta_3$ \\
			\midrule
			\multirow{6}{*}{\rotatebox[origin=l]{90}{\parbox[c]{2cm}{\centering \textbf{\textsc{Day}}}}}
			 & \textsc{Monodepth RGB \cite{Godard2017}} & 12.74  &  8.43  &  75.11  & \underline{ 90.18 } & \textbf{ 94.81 } & 14.04  &  9.10  &  72.70  &  88.43  & \underline{ 94.32 } & 14.67  &  10.64  &  63.49  &  82.90  & \underline{ 91.89 } & 13.17  &  8.73  &  71.56  &  89.06  & \textbf{ 94.81 } \\ 
 & \textsc{Sparse-to-Dense \cite{ma2018sparse}} & 13.66  &  9.85  &  54.20  &  82.42  &  91.47  & 14.23  &  10.66  &  49.75  &  79.62  &  90.10  & 18.50  &  15.35  &  37.04  &  64.67  &  78.25  & 13.42  &  9.81  &  53.12  &  82.29  &  92.04  \\ 
 & \textsc{Packnet-Slim G \cite{guizilini20203d}} & 16.46  &  11.62  &  56.91  &  78.43  &  88.48  & 16.95  &  11.80  &  59.09  &  78.80  &  88.81  & 17.01  &  12.09  &  54.93  &  76.37  &  88.89  & 15.30  &  10.33  &  62.22  &  82.29  &  90.65  \\ 
 & \textsc{Monodepth2 G \cite{godard2019digging}} & 13.26  &  7.40  &  78.59  &  88.95  &  92.77  & 18.17  &  10.43  &  72.91  &  83.20  &  89.27  & 15.56  &  8.72  & \underline{ 76.79 } &  85.38  &  90.68  & 12.84  &  7.12  & \textbf{ 80.04 } & \underline{ 89.34 } &  93.13  \\ 
 & \textsc{Gated2Depth \cite{gated2depth2019}} & \underline{11.48 } & \underline{ 6.60 } & \underline{ 79.17 } &  87.38  &  91.58  & \underline{11.28 } & \underline{ 6.63 } & \underline{ 81.20 } & \underline{ 88.66 } &  92.56  & \underline{11.86 } & \underline{ 7.85 } &  71.72  & \underline{ 87.10 } &  91.70  & \underline{11.28 } & \underline{ 6.61 } &  78.87  &  87.93  &  92.50  \\ 
 & \textsc{\textbf{Gated2Gated}} & \textbf{11.15 } & \textbf{ 6.31 } & \textbf{ 80.82 } & \textbf{ 90.48 } & \underline{ 93.97 } & \textbf{10.70 } & \textbf{ 6.01 } & \textbf{ 84.71 } & \textbf{ 91.52 } & \textbf{ 94.65 } & \textbf{11.09 } & \textbf{ 6.86 } & \textbf{ 81.09 } & \textbf{ 91.43 } & \textbf{ 94.47 } & \textbf{10.97 } & \textbf{ 6.28 } & \underline{ 80.01 } & \textbf{ 91.12 } & \underline{ 94.63 } \\

			\midrule
			\multirow{6}{*}{\rotatebox[origin=l]{90}{\parbox[c]{2cm}{\centering \textbf{\textsc{Night}}}}}
			 & \textsc{Monodepth RGB \cite{Godard2017}} & 13.78  &  8.92  &  72.63  &  88.48  & \underline{ 93.37 } & 13.30  &  8.75  &  72.52  &  88.88  & \textbf{ 94.45 } & 16.31  &  10.99  &  69.15  &  85.92  & \textbf{ 91.33 } & 15.28  &  9.89  &  68.88  &  86.86  & \textbf{ 93.44 } \\ 
 & \textsc{Sparse-to-Dense \cite{ma2018sparse}} & 14.43  &  10.40  &  50.32  &  78.66  &  89.58  & 13.92  &  10.01  &  51.88  &  80.41  &  90.70  & 16.54  &  12.07  &  47.15  &  73.45  &  84.36  & 14.08  &  10.05  &  52.91  &  81.03  &  90.85  \\ 
 & \textsc{Packnet-Slim G \cite{guizilini20203d}} & 15.81  &  11.11  &  59.80  &  79.52  &  88.65  & 16.01  &  11.23  &  58.44  &  80.60  &  89.57  & 17.49  &  12.60  &  57.72  &  77.77  &  87.26  & 16.47  &  11.40  &  60.17  &  79.55  &  88.62  \\ 
 & \textsc{Monodepth2 G \cite{godard2019digging}} & 14.52  &  8.30  &  74.45  &  85.41  &  89.73  & 14.21  &  8.29  &  74.56  &  85.06  &  91.01  & 18.33  &  11.88  &  66.61  &  79.25  &  84.48  & 15.11  &  8.46  &  76.15  &  86.38  &  90.52  \\ 
 & \textsc{Gated2Depth \cite{gated2depth2019}} & \textbf{10.06 } & \textbf{ 5.17 } & \textbf{ 84.81 } & \textbf{ 90.59 } & \textbf{ 93.39 } & \textbf{9.94 } & \textbf{ 5.37 } & \textbf{ 81.95 } & \textbf{ 89.80 } & \underline{ 93.63 } & \textbf{12.51 } & \textbf{ 7.72 } & \textbf{ 76.90 } & \underline{ 86.59 } & \underline{ 90.81 } & \textbf{10.70 } & \textbf{ 5.81 } & \textbf{ 81.81 } & \underline{ 89.45 } &  93.02  \\ 
 & \textsc{\textbf{Gated2Gated}} & \underline{11.69 } & \underline{ 6.74 } & \underline{ 80.25 } & \underline{ 89.58 } &  92.83  & \underline{11.29 } & \underline{ 6.46 } & \underline{ 79.39 } & \underline{ 89.31 } &  93.17  & \underline{13.52 } & \underline{ 8.69 } & \underline{ 76.43 } & \textbf{ 86.70 } &  90.61  & \underline{11.91 } & \underline{ 6.80 } & \underline{ 80.76 } & \textbf{ 90.09 } & \underline{ 93.31 } \\

			\bottomrule	
		\end{tabular}
	}
	\vspace*{-3pt}
	\caption{\label{tab:adv_weather_results} Evaluation of the proposed Gated2Gated framework and state-of-the-art-methods on adverse weather scenes. All metrics are evaluated in bins of approximately 7m to weight all distances equally. G indicates training and evaluation on gated images. Best results in each category are in \textbf{bold} and second best are \underline{underlined}.
		\vspace*{-4mm}
	}
\end{table*}

\vspace{-10pt}
\section{Dataset}
In order to train our proposed models, we collected 
1835 video sequences, which comprise about 130,000 frames
(note that previous gated imaging datasets~\cite{gated2depth2019,Bijelic_2020_STF} 
do not include sequential data). 
Each time history is centered around one of the 13,000 middle frames 
and provides a temporal history of \unit[1]{second} at 
a sampling rate of \unit[10]{Hz}. The center frames are 
pre-selected by human annotators depending on the scene 
of interest from an underlying data distribution covering 
diverse winter road scenes collected in Northern Europe 
 following the settings proposed in~\cite{Bijelic_2020_STF}. 
Please see the supplemental material for additional information.

We evaluate our proposed method on the open-source 
Gated2Depth~\cite{gated2depth2019} and Seeing Through
Fog~\cite{Bijelic_2020_STF} datasets. While the first dataset 
contains a large variety of day and night captured images, 
the second contains diverse cluttered recordings in light fog, 
dense fog and snowfall conditions, which allow us to evaluate 
the performance of our model in harsh weather conditions and 
scenarios where obtaining ground truth data is difficult. 
In the last case, to filter out clutter from LiDAR ground truth, 
we use the DROR filtering algorithm~\cite{DRORFilter}, 
removing \unit[8.2]{\%} LiDAR points.

\vspace{-0.12cm}
\section{Experiments}
\vspace{-0.12cm}
In this section, we evaluate the proposed method by comparing it against the existing
depth estimation approaches based on monocular 
gated, RGB images, and stereo RGB images; 
in both supervised and self-supervised training settings. 
\subsection{Implementation Details}
\vspace{-0.12cm}
Although the proposed approach is not limited to a specific architecture, we estimate $f_\mathbf{r}$ depth maps with the
PackNet~\cite{guizilini20203d} network architecture. 
To estimate ambient $f_{\mathbf{\Lambda\alpha}}$ and  albedo $\ahat_t$, we use a UNet-based\cite{ronneberger2015u} 
architecture with a single encoder and two decoder heads.
Pose transformation $f_{\mathbf{t\rightarrow n}}$ is 
learned through the model introduced by Zhou \textit{et al.}\cite{Zhou2017}, without the explainability mask.
The joint neural networks are implemented using PyTorch\cite{paszke2017automatic}. We define a 
$512 \times 1024$ gated image input resolution and trained 
the model on an NVIDIA A100 GPU with a batch size of four. 
As optimizer we use ADAM \cite{kingma2014adam} with
$\beta_1\,=\,0.9$ and $\beta_2\,=\,0.9999$ and 
learning rate of $10^{-4}$. %
In total, the model is trained for 30 epochs, 
with first 10 epochs used to train the depth
$f_\mathbf{r}$ and pose $f_\mathbf{t \rightarrow n}$ prediction networks using temporal consistency only, and the last 20 epochs to jointly train ambient+albedo $f_\mathbf{\Lambda\alpha}$ alongside depth and pose. 
For valid pixel masks, we set $\gamma=0.98,\ \theta=0.04$ and $c=0.995$. 

\begin{table*}[!t]
	\centering
	\setlength{\tabcolsep}{1pt} 
	\resizebox{1.01\textwidth}{!}{
		\begin{tabular}{@{}>{\centering\arraybackslash}m{0.25cm} cccc>{\centering\arraybackslash}m{3.2cm} c@{}}
			\multirow{1}{*}{\hspace{-0.05cm}\rotatebox[origin=l]{90}{\parbox[c]{0.6cm}{\centering \small $\mathbf{Z}_t$}}} &
			\multirow{4}{*}{\includegraphics[height=6.0cm]{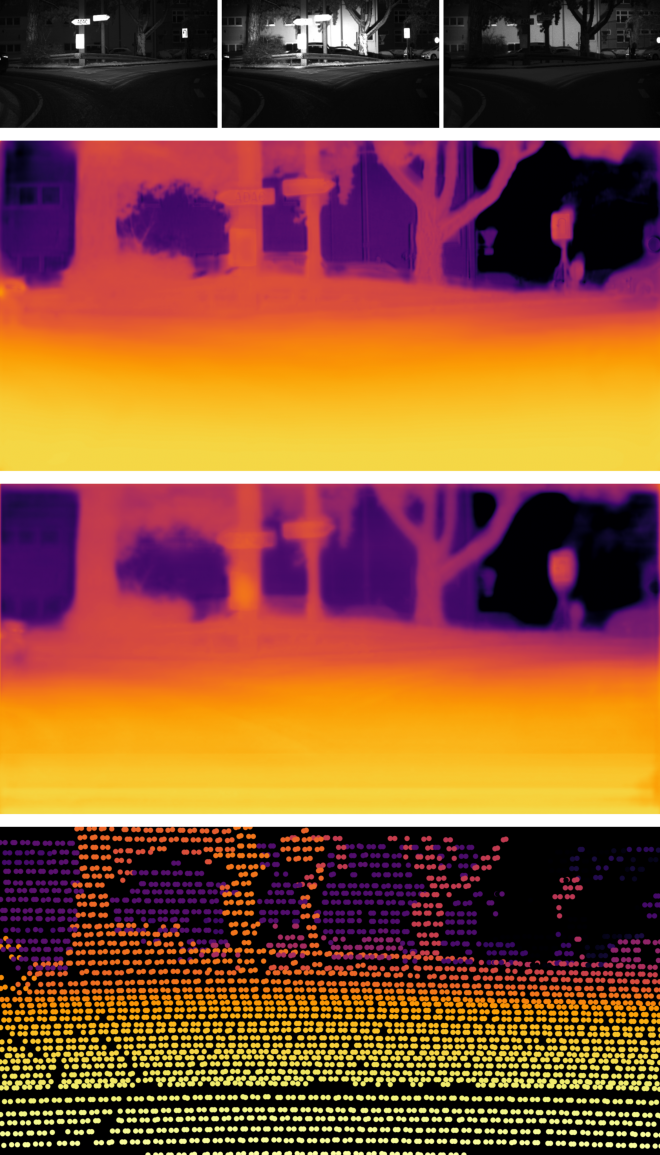}} &
			\multirow{4}{*}{\includegraphics[height=6.0cm]{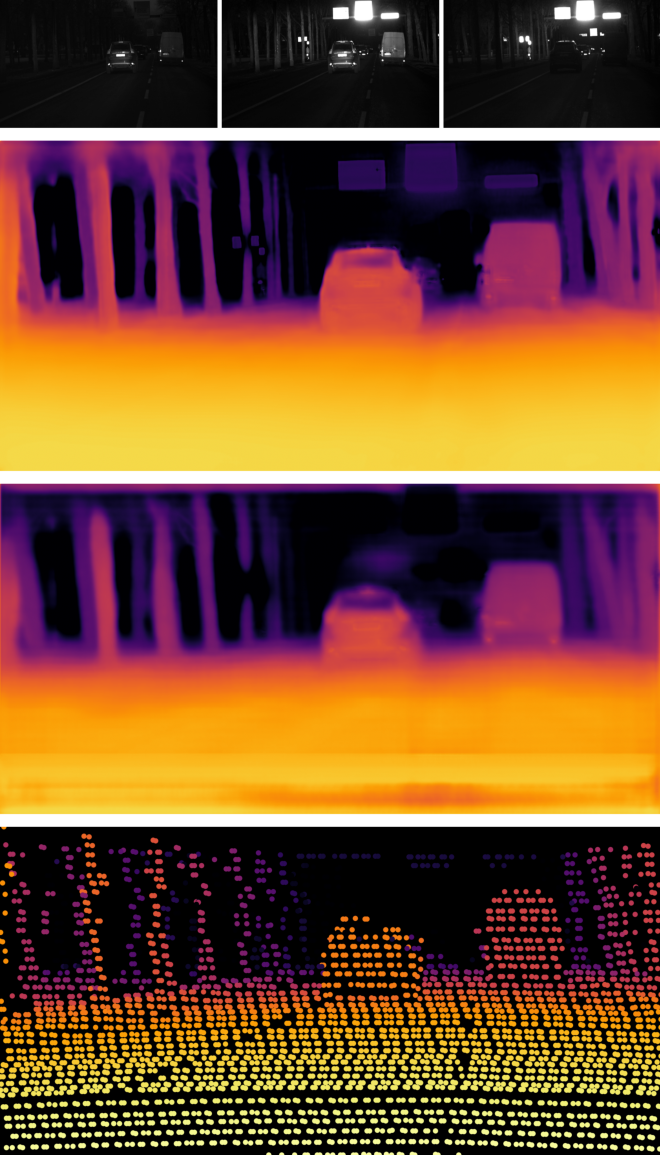}}  &
			\multirow{4}{*}{\includegraphics[height=6.0cm]{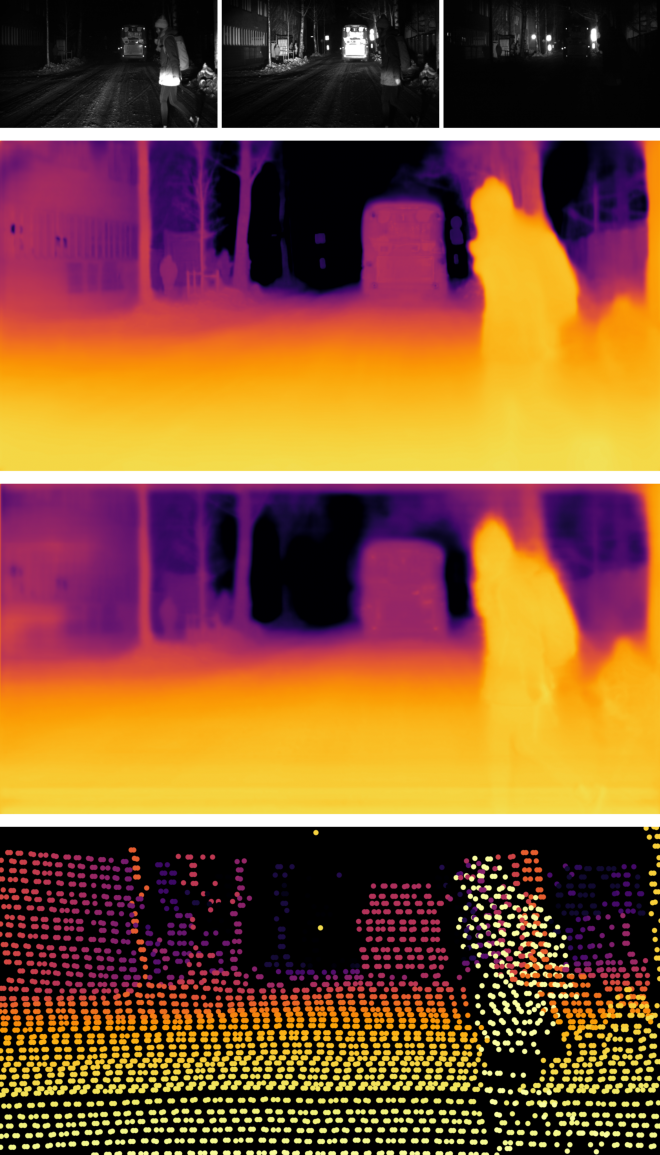}} &
			\multirow{4}{*}{\includegraphics[height=6.0cm]{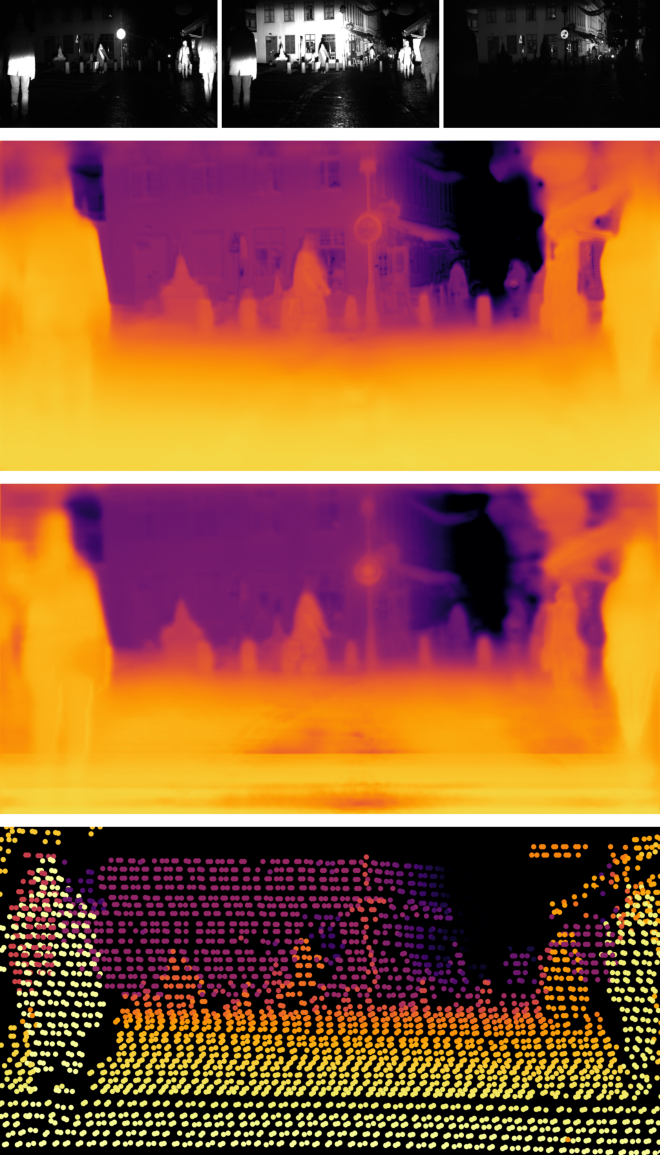}}  &
			\multirow{4}{*}{\includegraphics[height=6.0cm]{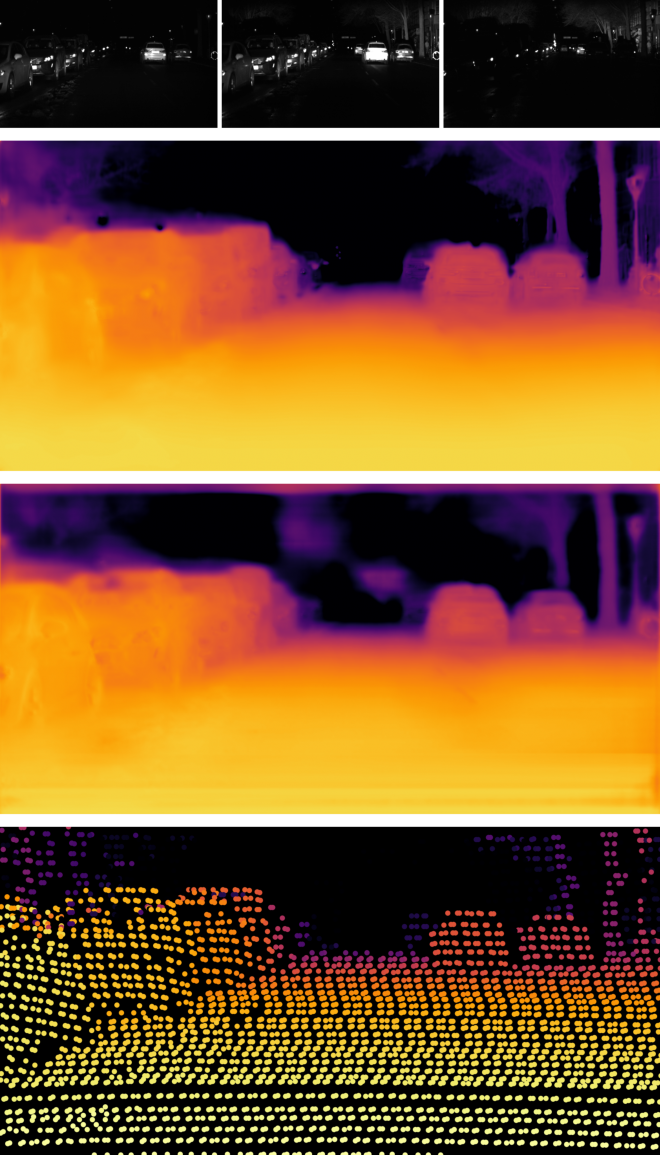}}  & [m] 	\\
			\multirow{1}{*}{\rotatebox[origin=l]{90}{\parbox[c]{2.3cm}{\centering \small \textbf{G2G}}}} & & & & & & \multirow{3}{*}[-.5mm]{\includegraphics[height=5.5cm]{imgs/colorbar/colorbar_g2d_vs_g2g.pdf}}\\
			\multirow{1}{*}{\rotatebox[origin=l]{90}{\parbox[c]{5.1cm}{\centering \small G2D}}} & & & & & &\\
			\multirow{1}{*}{\rotatebox[origin=l]{90}{\parbox[c]{7.75cm}{\centering \small LiDAR}}} & & & & & & \\ 
			\vspace*{9.1cm}
			\multirow{1}{*}[4.4cm]{\hspace{-0.05cm}\rotatebox[origin=l]{90}{\centering \small $\mathbf{Z}_t$}} &
			\multirow{4}{*}{\includegraphics[height=6.0cm]{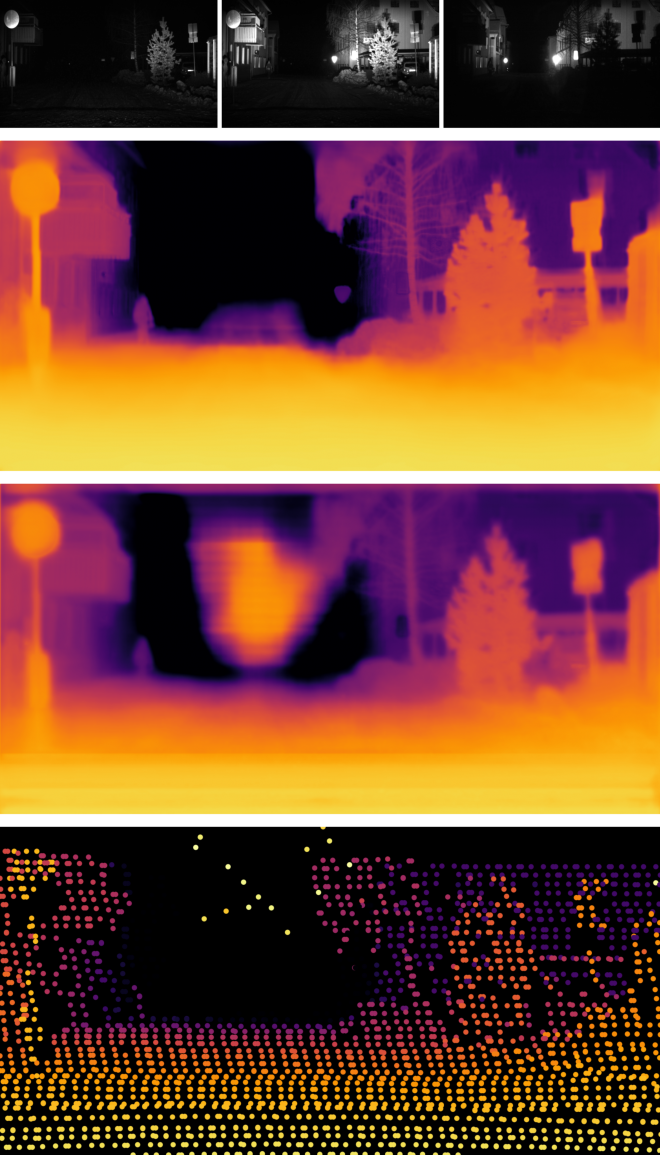}}  &
			\multirow{4}{*}{\includegraphics[height=6.0cm]{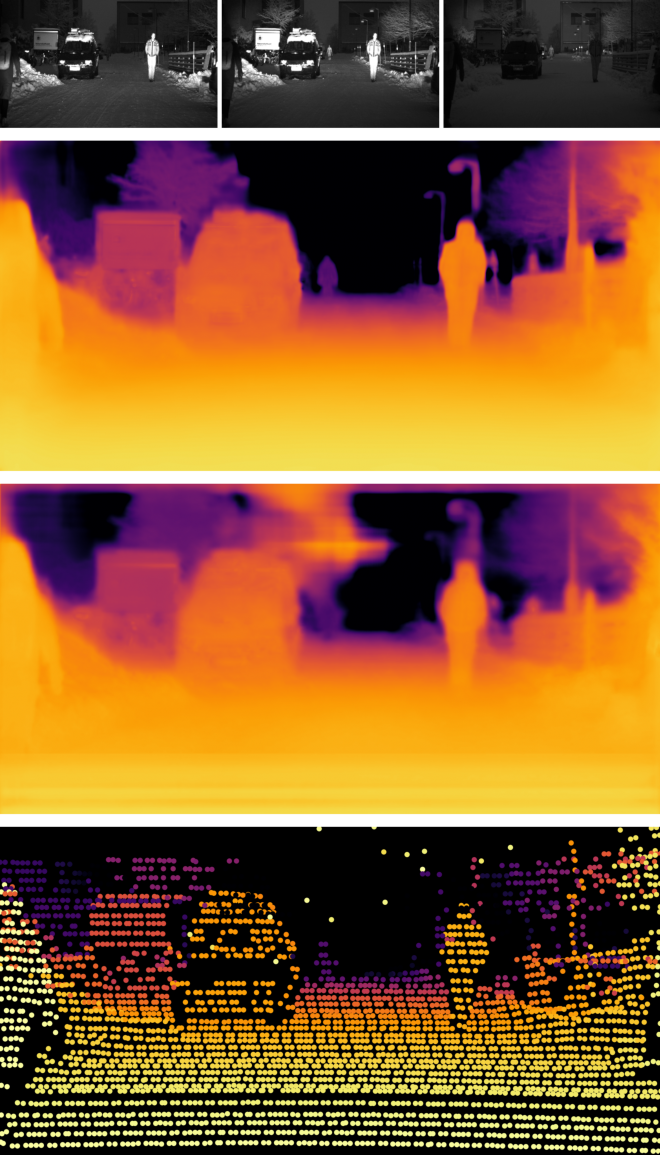}} &
			\multirow{4}{*}{\includegraphics[height=6.0cm]{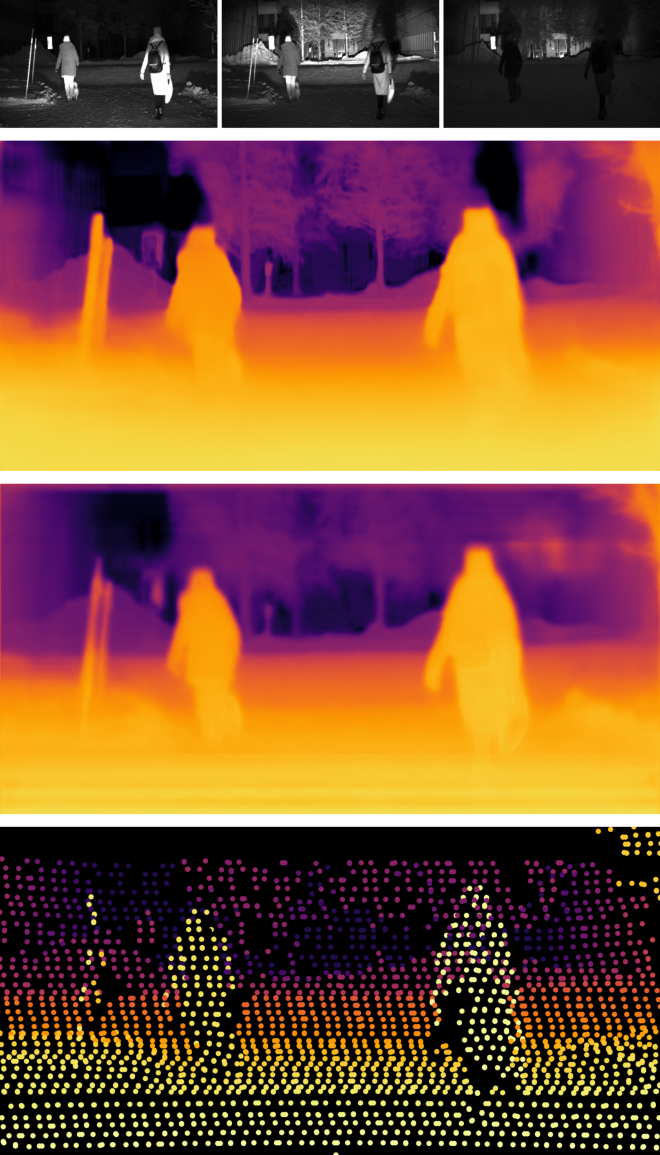}} &
			\multirow{4}{*}{\includegraphics[height=6.0cm]{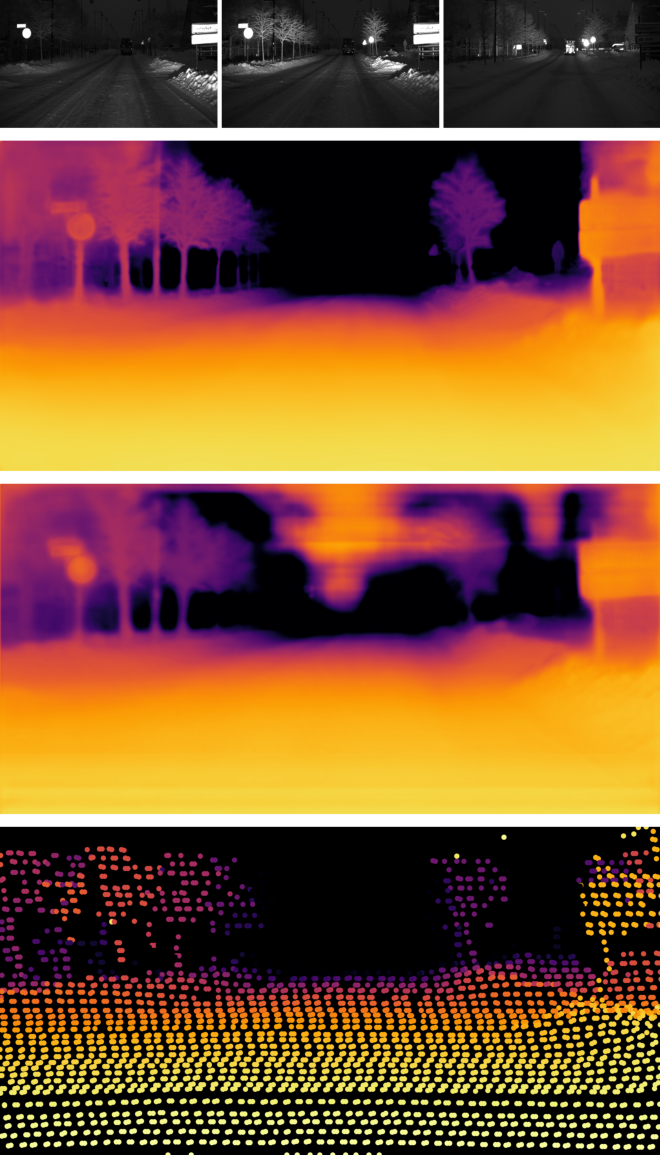}} &
			\multirow{4}{*}{\includegraphics[height=6.0cm]{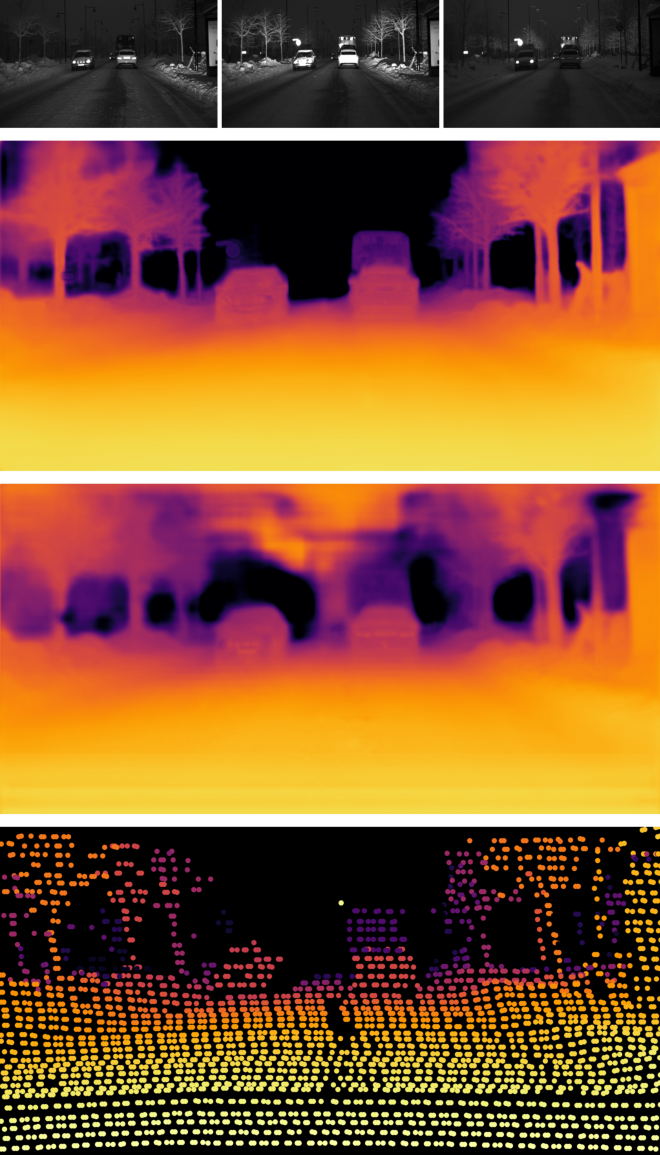}} & [m] 	\\
			\multirow{1}{*}[3.75cm]{\rotatebox[origin=l]{90}{\centering \small \textbf{G2G}}} & & & & & & 
			\multirow{3}{*}[4.5cm]{\hspace{-2.25pt} \includegraphics[height=5.5cm]{imgs/colorbar/colorbar_g2d_vs_g2g.pdf}}\\
			\multirow{1}{*}[2.3cm]{\rotatebox{90}{\small G2D}} & & & & & & \\
			\multirow{1}{*}[1.1cm]{\rotatebox{90}{\small LiDAR}} & & & &  &  & \\  
			
	\end{tabular}}
	\vspace*{-0.35cm }
	\captionof{figure}{The top row for each example shows the gated image ${\mathbf{Z}_{t}}$ with three gated slices ($\Z^{0}_{t}$, $\Z^{1}_{t}$, $\Z^{2}_{t}$) for each capture, second row shows depth maps predicted by the \textbf{Gated2Gated (G2G)} -- self-supervised method, third row shows depth predictions from Gated2Depth (G2D)\cite{gated2depth2019}  -- supervised, and the bottom row shows LiDAR point cloud in gated view.}
	\label{fig:g2g_vs_g2d}
\end{table*}

\subsection{Assessment}
\paragraph{Experimental Setup.}
We compare the performance of the proposed Gated2Gated method against state-of-the-art supervised and self-supervised depth estimation methods. As supervised approaches, we compare against gated depth estimation \cite{gated2depth2019,adam2017bayesian}, LiDAR depth completion \cite{ma2018sparse} and stereo vision \cite{chang2018pyramid, Hirschmuller2008}. For comparison with unsupervised methods, we consider stereo \cite{godard2017unsupervised} and temporal based self-learning approaches \cite{godard2019digging, guizilini20203d}. Since the self-supervised baseline methods 
do not provide absolute depth predictions, we follow previous
works \cite{Zhou2017, godard2019digging, guizilini20203d} and
scale the estimated depth maps with median ground-truth LiDAR
information. For completeness, we train and evaluate these
approaches both on gated and on RGB input images. 
All supervised methods and \cite{godard2017unsupervised} are
trained on the Gated2Depth training set, 
where LiDAR labels are available. The self-supervised
monocular approaches are trained on the proposed temporal
dataset. For further training details, we refer the readers to the supplemental document. \\
Following \cite{eigen2014depth}, we evaluate using the metrics
RMSE, MAE, ARD and $\delta_i < 1.25i$ for $i \in {1,2,3}$.
These metrics are computed for distances between 3m and 80m,
limited by the maximum LiDAR distance. To evaluate the long
range influence in adverse weather we rely 
on \unit[7]{m} bins. We apply binned metrics for 
cluttered data, since in adverse conditions LiDAR
measurements are not equally distributed over  
distances~\cite{gruber2019pixel}. %

\vspace{-1.5eM}
\paragraph{Evaluation on Clear Data - Gated2Depth
dataset.}
Table \ref{tab:results_g2d} reports a quantitative 
comparison of our proposed Gated2Gated method and 
other state-of-the-art methods on the test set of the
Gated2Depth dataset \cite{gated2depth2019}. 
Our model outperforms all other self-supervised methods
\cite{guizilini20203d, godard2019digging, godard2017unsupervised}, and even 
temporal approaches \cite{guizilini20203d, godard2019digging} that use 
LiDAR ground truth for depth scaling. 
Our proposed method also outperforms 
stereo \cite{Hirschmuller2008, chang2018pyramid} and 
Bayesian-based gated depth estimation 
methods \cite{adam2017bayesian}. Among the supervised methods,
only Sparse-to-Dense \cite{ma2018sparse} and Gated2Depth
\cite{gated2depth2019} obtain better results. 
We note, however, that Sparse-to-Dense \cite{ma2018sparse} 
relies on sparse ground truth depth inputs from LiDAR sensors.
Figure \ref{fig:comp_ref_methods} qualitatively compares 
our method to baseline methods for depth estimation. 
In Figure \ref{fig:comp_ref_methods_snow}, our method accurately 
shows all scene objects located at a farther distance -- car, traffic
signs and two pedestrians in adverse weather 
conditions like snowfall, whereas RGB-based and LiDAR-based 
methods deliver only poor depth predictions. While
Gated2Depth\cite{gated2depth2019} is also able to 
recover the scene elements, the generated depth map 
shows artifacts at a farther distance, unlike our method. 
Similarly in Figure \ref{fig:comp_ref_methods_day}, our 
method generates accurate depth maps whereas all other 
methods fail here. Figure \ref{fig:g2g_vs_g2d} shows the qualitative comparison of the
supervised Gated2Depth \cite{gated2depth2019} and the proposed  
self-supervised approach. While the performance metrics
for Gated2Depth are on par with the proposed method, the qualitative comparison shows 
that our method predicts much finer grain details and sharper 
object contours in depth maps. 
Furthermore, the proposed method generalizes better 
to far distances: Gated2Depth often estimates far distances and sky  
as close regions. \\

\vspace{-2.5eM}
\paragraph{Ablation Study.}
To evaluate the individual contributions, we perform an ablation study reported in Table~\ref{tab:results_g2d}. For the base method without valid masks, we can see that the performance is the worst and improves when adding $b$ or $v$ for the RMSE score by \unit[9.1]{\%}. Training with both masks is mutually beneficial and provides a significant boost to overall performance metrics. We provide additional ablation experiments in the Supplemental Document.

\paragraph{Evaluation on Adverse Weather Scenes -- Seeing Through Fog dataset.}
We also evaluate the proposed method in adverse weather,
adopting the same test splits provided in
\cite{Bijelic_2020_STF}. The performance is measured in 
binned metrics to weight all distances equally. 
Table \ref{tab:adv_weather_results} shows the 
quantitative results of the Gated2Gated method and
state-of-the-art methods. We note that absolute metrics 
may improve in adverse weather conditions, 
as the number and range of ground-truth LiDAR points 
decreases with worse weather conditions. 
We validate that Gated2Gated achieves robust 
performance overall weather conditions. 
In contrast, Monodepth2 and Sparse-to-Dense 
struggle to maintain performance in  
adverse weather. Since Sparse-to-Dense uses 
LiDAR points as additional inputs, wrong 
depth measurements from backscatter negatively impact 
the predicted depth maps. Furthermore,  
Table \ref{tab:adv_weather_results} validates that 
the Gated2Gated approach performs on par with 
Gated2Depth and for daytime and in harsh weather scenarios 
our approach even outperforms Gated2Depth.
These results highlight the generalization capabilities 
of Gated2Gated over a wide range of distances 
and weather conditions.

\section{Conclusion}
We introduce Gated2Gated as a method that learns to estimate depth from gated images in a self-supervised fashion -- only by observing gated video sequences. The proposed method exploits cyclic measurement and temporal consistency cues as training signals. This method can resolve monocular scale ambiguity by relying on gated illumination profiles and shadow/multi-path reflection via multi-view observations. To train the self-supervised method, we create a novel gated video dataset containing 130,000 frames from 1835 sequences. We validate Gated2Gated in extensive real-world experimentation, where it outperforms fully supervised methods by up to \unit[1.25]{m} ($\downarrow$ 11.25\%) on the RMSE metric in daytime adverse weather and by at least \unit[1.2]{m} ($\downarrow$ 9.73\%) independently of adverse weather for the RMSE metric. In the future, we plan to add wide-baseline active stereo cues to our self-supervised method by using two synchronized gated imagers.

{\small
   \bibliographystyle{ieee_fullname}
   \bibliography{egbib}
}

\end{document}